\documentclass[journal]{IEEEtran}
\usepackage{amsmath,amsfonts}
\usepackage{array}
\usepackage[caption=false,font=footnotesize,labelfont=rm,textfont=rm]{subfig}
\usepackage{textcomp}
\usepackage{stfloats}
\usepackage{url}
\usepackage{verbatim}
\usepackage{graphicx}
\usepackage{cite}
\usepackage{multirow}
\usepackage{orcidlink}
\usepackage{tikz}
\usepackage[ruled,linesnumbered]{algorithm2e}
\usepackage{booktabs}
\usepackage{flushend}

\hypersetup{
  colorlinks=true,
  linkcolor=black,
  citecolor=black
}
\usepackage{hyperref}
\usepackage{cleveref}

\begin{document}

\title{MoCTEFuse: Illumination-Gated Mixture of Chiral Transformer Experts for Multi-Level Infrared and Visible Image Fusion}

\author{Jinfu Li$^{\orcidlink{0009-0008-6013-5675}}$,
        Hong Song$^{\orcidlink{0000-0002-3171-2604}}$,
        Jianghan Xia$^{\orcidlink{0009-0008-8826-0646}}$,
        Yucong Lin$^{\orcidlink{0000-0002-9039-0318}}$,
        Ting Wang$^{\orcidlink{0009-0000-6638-8130}}$,
        Long Shao$^{\orcidlink{0009-0003-2533-631X}}$,
        Jingfan Fan$^{\orcidlink{0000-0003-4857-6490}}$,
        and Jian Yang$^{\orcidlink{0000-0003-1250-6319}}$
\thanks{Corresponding authors: Hong Song and Jian Yang (E-mail: songhong@bit.edu.cn; jyang@bit.edu.cn)}
\thanks{Jinfu Li, Yucong Lin, Long Shao, Jinfan Fan and Jian Yang are with the School of Optics and Photonics, Beijing Institute of Technology, Beijing 100081, China.}
\thanks{Hong Song, Jianghan Xia, and Ting Wang are with the School of Computer Science and Technology, Beijing Institute of Technology, Beijing 100081,
China.}}


\maketitle

\begin{abstract}
  While illumination changes inevitably affect the quality of infrared and visible image fusion, many outstanding methods still ignore this factor and directly merge the information from source images, leading to modality bias in the fused results. To this end, we propose a dynamic multi-level image fusion network called MoCTEFuse, which applies an illumination-gated Mixture of Chiral Transformer Experts (MoCTE) to adaptively preserve texture details and object contrasts in balance. MoCTE consists of high- and low-illumination expert subnetworks, each built upon the Chiral Transformer Fusion Block (CTFB). Guided by the illumination gating signals, CTFB dynamically switches between the primary and auxiliary modalities as well as assigning them corresponding weights with its asymmetric cross-attention mechanism. Meanwhile, it is stacked at multiple stages to progressively aggregate and refine modality-specific and cross-modality information. To facilitate robust training, we propose a competitive loss function that integrates illumination distributions with three levels of sub-loss terms. Extensive experiments conducted on the DroneVehicle, MSRS, TNO and RoadScene datasets show MoCTEFuse's superior fusion performance. Finally, it achieves the best detection mean Average Precision (mAP) of 70.93\% on the MFNet dataset and 45.14\% on the DroneVehicle dataset. The code and model are released at \url{https://github.com/Bitlijinfu/MoCTEFuse}.
\end{abstract}

\begin{IEEEkeywords}
Infrared and visible image fusion, illumination-aware gate, mixture of chiral Transformer experts, asymmetric cross-attention, competitive loss function.
\end{IEEEkeywords}

\section{Introduction} \label{Introduction}
\IEEEPARstart{I}{nfrared} (IR) and visible (VI) image fusion aims to generate fused images that are not only visually appealing but also practical for downstream tasks. This technology attracts the attention of fields that need all-weather and around-the-clock perception, \textit{e.g.}, remote sensing \cite{2024OAFA}, autonomous driving \cite{huang2023multi}, and bio-robot\cite{2023robot}. Generally speaking, IR images, captured in the infrared spectrum, reflect the distribution of thermal radiation, so they are insensitive to surrounding environments. VI images, recorded in the visible light spectrum, have higher spatial resolution and more true-to-life appearances. However, the quality of VI images varies greatly under multi-illumination conditions, and some may even malfunction. If this factor is not properly treated, it will degrade the fusion performance. Therefore, a robust framework and model are necessary to mitigate the impact of illumination on rendering fused images.

\begin{figure*}[!htbp]
  \centering
  \includegraphics[width=\linewidth]{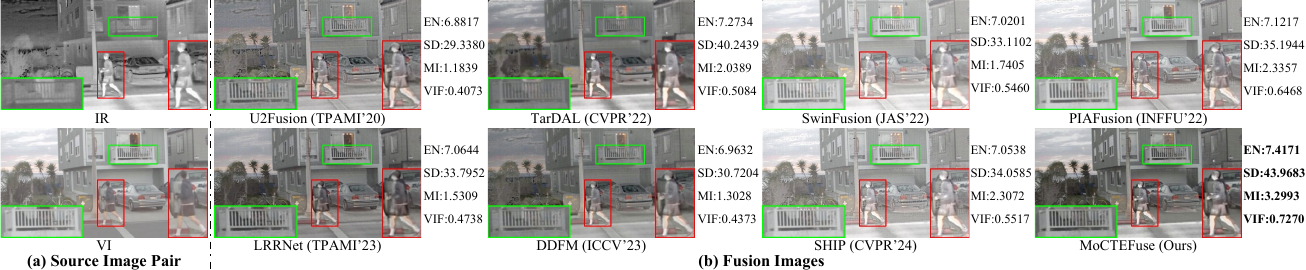}
  \caption{Comparisons of eight representative state-of-the-art methods. Thanks to its modality-adaptive dynamic fusion, MoCTEFuse has both subjective and objective advantages over the other methods, \textit{e.g.}, U2Fusion\cite{xu2020u2fusion}, TarDAL\cite{2022TarDAL}, SwinFusion\cite{2022SwinFusion}, PIAFusion\cite{tang2022piafusion}, LRRNet\cite{li2023lrrnet}, DDFM\cite{2023DDFM}, and SHIP\cite{2024ship}.}
  \label{fig:motivation}
\end{figure*}

In the past decades, many works have emerged for IR and VI image fusion, which can be classified into conventional and deep learning methods. Conventional methods usually extract features through mathematical manipulations, \textit{e.g.}, wavelet transform \cite{hill2016perceptual}, contourlet transform \cite{2010MCT}, sparse representation \cite{2020MDLatLRR}, subspace clustering \cite{cvejic2007region}. These learned features are then merged in spatial or transform domains by different measurements, \textit{e.g.}, coefficient, window, and region \cite{liu2015general,dogra2017multi,zhu2017fusion}. These methods sometimes offer good interpretability and outcomes, but over-reliance on manual features and predefined rules restricts their use in complex and unknown environments. As desirable alternatives to conventional methods, deep learning methods leverage deep neural networks to automatically capture and integrate features. They are roughly categorized into those based on convolutional neural networks (CNNs) \cite{xu2020u2fusion,2024ship}, generative adversarial networks (GANs) \cite{2022TarDAL,Wang_CrossFuse}, transformers \cite{2022SwinFusion,LImixfuse}, diffusion \cite{2023DDFM}, and hybrid networks \cite{tang2022ydtr,li2023lrrnet}. However, most of them tend to directly combine and recover details and contrasts without considering light changes in real scenes. \autoref{fig:motivation} demonstrates that they fail to maintain essential information from the respective modality.

Recently, researchers followed a preprocessing-to-fusion pipeline to tackle light disturbances. For low-light VI images, Tong et al. \cite{2019Tong} perform brightening and denoising procedures using Retinex theory and Rolling Guidance Filter. Zhang et al. \cite{zhang2024evfusion} realize simultaneous enhancement of intensity and color information. Yang et al. \cite{yang2024iaifnet} regulate the incident illumination maps to refine texture details. Besides, Luo et al. \cite{LUO2024FVC} combine Taylor approximation with visual compensation to add the visual fidelity of IR and VI images. While some images benefit from this fixed paradigm, others suffer from over-processing or under-correction, weakening the overall fusion quality. Given this limitation, a few studies began exploring a flexible and adaptable paradigm. In \cite{tang2022piafusion}, PIAFusion utilizes a daytime/nighttime classification network to predict the illumination probability and embed it into the intensity loss. MoEFusion~\cite{cao2023moefusion} is the first sample-adaptive framework with a mixture of local-to-global experts. Despite these promising efforts, they were not entirely flawless. Firstly, PIAFusion often miss critical cues that enrich the visual experience due to its single illumination loss term. As shown in \autoref{fig:motivation}, the model disregards certain details and contrast variations in the IR image, yielding outputs that lack depth and realism. Secondly, in MoEFusion, the gating network's input includes illumination-invariant features from IR images, affecting the prediction accuracy of the gate. The segregation of local and global learning also hinders balanced capture of local details and global structures, further limiting the performance.

\par{
  Since high-illumination conditions mainly capture information in the VI modality, and low-illumination conditions rely more on the IR modality, most existing algorithms fail to adjust the modality priority according to illumination conditions. Hence, the fused images often favor one modality over the other. To tackle this issue, we introduce chiral transformer experts for dynamic switching between primary and auxiliary modalities. \autoref{fig:motivation} shows that our method mitigates modality bias and achieves modality-adaptive fusion, delivering superior performance. Concretely, we propose MoCTEFuse, a dynamic hierarchical image fusion framework with an illumination-gated Mixture of Chiral Transformer Experts (MoCTE). It comprises high-illumination (HI-MoCTE) and low-illumination (LI-MoCTE) subnetworks guided by an illumination-aware gate. The core of HI-MoCTE and LI-MoCTE is to employ the chiral transformer fusion block (CTFB) to iteratively model inter- and intra-modal relationships across various hierarchy levels. Within each CTFB, an asymmetric cross-attention mechanism is devised to dynamically adjust its focus based on the modality-specific importance. In a word, the features from the primary modality serve as the queries, while the features from the auxiliary modality function as supplements for the keys and values. This adaptability allows the model to emphasize the most informative features while suppressing irrelevant information. To facilitate network training, we introduce a competitive loss function that integrates illumination, structural similarity, intensity, and gradient terms. The main contributions of our work are summarized as follows:
}

\begin{enumerate}
\item We propose a dynamic multi-level framework for reliable infrared and visible image fusion. The proposed framework applies a novel illumination-gated MoCTE to effectively adapt to illumination changes.
\item We devise an asymmetric cross-attention-based CTFB to dynamically integrate valuable information from each modality and iteratively enhance it by modeling inter- and intra-modal relationships at various levels.
\item We define a competitive loss function for robust training, which unifies three sub-loss terms with illumination factors to promote modality-adaptive fusion.
\item Subjective and objective results on four benchmarks confirm the superiority of our framework over many excellent methods. Additionally, our method achieves the highest detection performance on two real scenes.
\end{enumerate}

\par{
  The rest content is organized as follows: \autoref{section2} briefly reviews related works. \autoref{section3} presents our method in detail. \autoref{section4} describes the experimental setup, results, and analyses. Finally, \autoref{section5} reports the conclusions.
}

\section{Related Work}\label{section2}
\subsection{Deep Learning-Based Image Fusion Networks}
\par{
  To the best of our knowledge, \cite{liu2018infrared} was the first to propose a CNN-based method for IR and VI image fusion. Densefuse \cite{RN25} pioneered the exploration of CNN in an auto-encoder architecture. It utilized the MS-COCO dataset to pretrain dense blocks in the encoder and adopted simple fusion rules (\textit{i.e.}, addition and L1-norm) to obtain the fused images. FusionGAN \cite{2019FusionGAN} introduced the first GAN-based method, where the generator aimed to produce fused images with infrared intensity, while the discriminator encouraged the incorporation of additional details from visible images. More recently, TarDAL \cite{2022TarDAL} and SeaFusion \cite{tang2022seafusion} cascaded image fusion tasks with object detection and semantic segmentation, respectively. This multi-task joint learning effectively addressed the problem of high fusion indices but low performance on downstream tasks. Furthermore, SwinFusion \cite{2022SwinFusion} designed intra-domain and inter-domain fusion units based on self-attention and cross-attention for general image fusion. The study clearly explained the importance of global information. CDDFuse \cite{zhao2023cddfuse} introduced a dual-branch Transformer-CNN structure accompanied by a correlated loss function. This structure decomposed source images into background and detail components based on high- and low-frequency information, respectively. However, the preceding works rarely considers the impact of real illumination variations on multi-modality representation, causing a loss of visible details in bright environments and reduced infrared contrasts in dim settings. We put forward a dynamic multi-level image fusion framework adapting to diverse samples. Guided by specialized experts, it generates fused images that strike a balance between details and contrasts.
}
\begin{figure*}[!htbp]
  \centering
  \includegraphics[width=\linewidth]{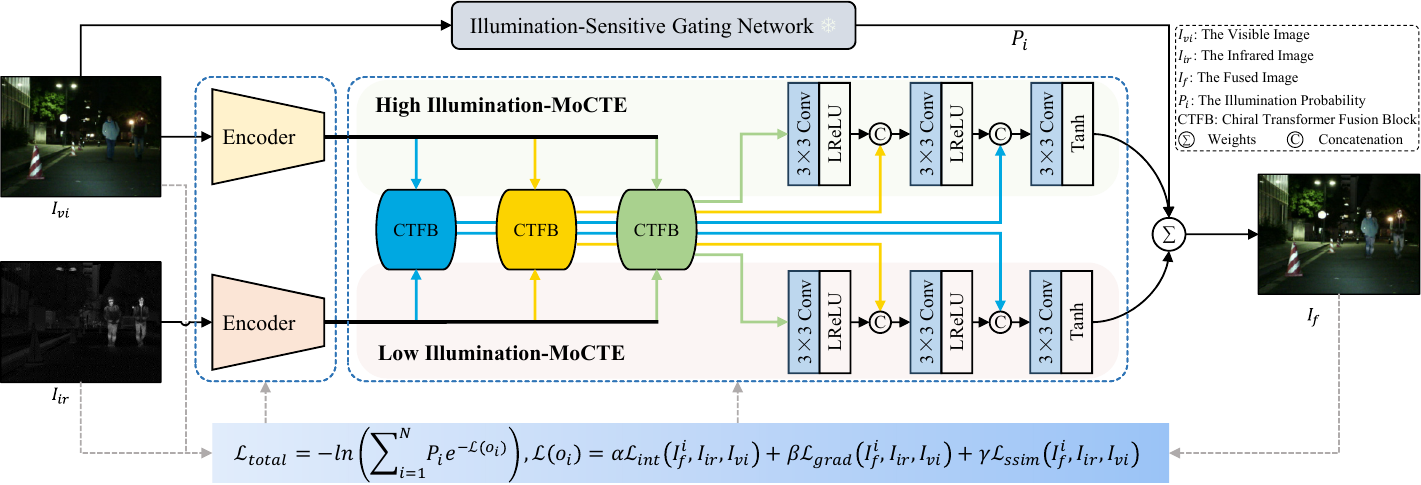}
  \caption{The framework of the proposed MoCTEFuse. The CTFB is designed to communicate and integrate cross-modality features according to illumination conditions. It operates at multiple hierarchical levels to progressively refine the extracted features in the fusion process. Additionally, the competitive learning loss function bridges the gap between illumination factors and network training.}
  \label{fig:MoCTEFuse}
\end{figure*}
\subsection{Illumination-Sensitive Vision Applications}
\par{
  Researchers developed numerous methods to handle illumination variations in practical vision systems. In \cite{MBNet-ECCV2020}, a modality balance network was present for detecting pedestrians across multispectral bands. This network employed an illumination-aware feature alignment module, which selectively leveraged complementary features based on illumination levels and aligned them adaptively. In contrast, \cite{2023CSIM} combined an illumination-invariant chromaticity space module with a YOLO-based detection framework to boost the effectiveness of intelligent traffic surveillance. Additionally, \cite{Guo_2020_CVPR} redefined the low-light image enhancement as image-specific curve estimation by using a lightweight network. \cite{Wang_2023_ICCV} designed two illumination-aware gamma correction modules to predict global and local adaptive correction factors.
}

\subsection{Mixture-of-Experts Networks}
\par{
  Mixture-of-Experts networks (MoEs) originated from pioneering research of \cite{6797059} and \cite{716791}. An important milestone is the application of MoEs in natural language processing (NLP) tasks. Numerous works \cite{2021Gshard, xue2022go} have demonstrated the excellent scalability of MoEs across varying input conditions. Currently, a growing number of research efforts \cite{NEURIPS2021_VMoE, chen2023mod, zhao2024removal} have extended MoEs into the field of computer vision. For example, Carlos et al. \cite{NEURIPS2021_VMoE} introduced a scalable and competitive Transformer model with MoEs for image recognition. Chen et al. \cite{chen2023mod} addressed the multitask learning challenge by incorporating MoEs into Transformer architecture, formalizing cooperation and specialization as expert and task matching processes. Zhao et al. \cite{zhao2024removal} employed a mixture of scale-aware experts to select desired features for interaction. These MoE-based methods merely focused on universal knowledge learning to manage experts, resulting in a lack of clear specialization. Instead, we extend MoEs to infrared and visible image fusion by creating an illumination-gated mixture of chiral transformer experts. This approach dynamically routed inputs to the most suitable expert, ensuring modality-adaptive fusion and achieving exceptional performance.
}

\section{Proposed Method} \label{section3}
\subsection{Framework Overview} \label{method:1}
\par{
  As shown in \autoref{fig:MoCTEFuse}, the proposed framework processes infrared image $I_{ir}$ and visible image $I_{vi}$ through two distinct encoders, tailored to capture thermal intensity and texture details, respectively. Firstly, it utilizes a convolutional layer with a $3\times3$ kernel and a leaky rectified linear unit (LReLU) activation function to extract shallow features from the input images. Secondly, the encoder incorporates Residual Transformer Block (RTB) and Residual Dense Blocks (RDB) to acquire deeper features \cite{Li_DCTNet}. Next, as the core of HI-MoCTE and LI-MoCTE, CTFB integrates the extracted features based on illumination conditions. Each subsequent CTFB layer builds upon the outputs of its predecessor, allowing the model to effectively enrich and maintain both modality-specific and cross-modality features. The detailed structure of CTFB is elaborated in \autoref{method:2}. Lastly, considering high- and low-illumination as equivalent to daytime and nighttime scenes \cite{MBNet-ECCV2020}, we develop an illumination-sensitive network to estimate the illumination of visible images. Its prediction probability $P_i$ is used to assign appropriate weights to HI-MoCTE and LI-MoCTE, with the weighted output being fused image $I_f$.
}
\subsection{Chiral Transformer Fusion Block} \label{method:2}
\begin{figure*}[!htbp]
  \centering
  \includegraphics[width=\linewidth]{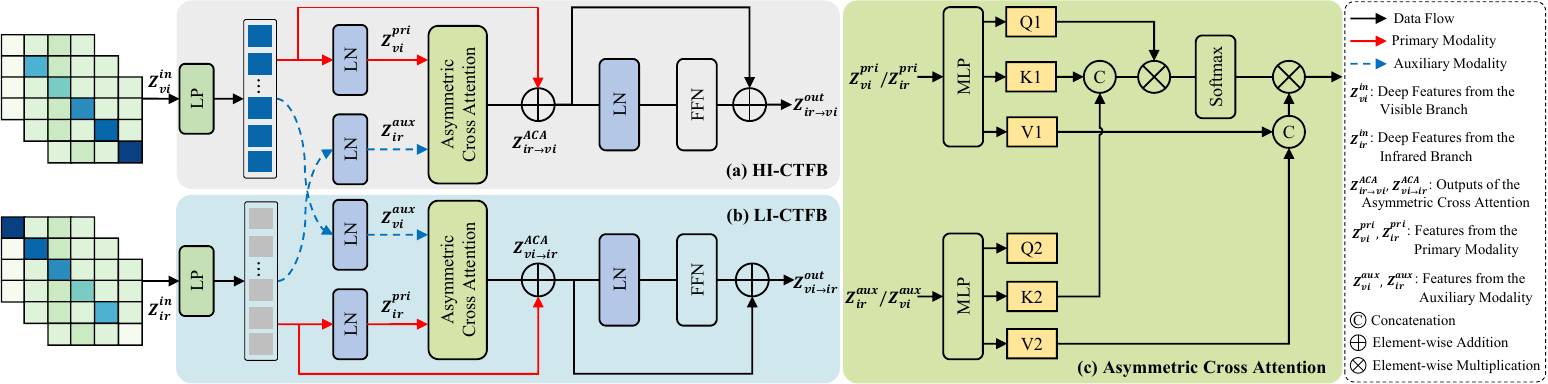}
  \caption{The structure of the CTFB. (a) HI-CTFB, where the visible modality is the primary modality, while the infrared modality functions as an auxiliary. (b) LI-CTFB, where the visible modality assumes an auxiliary role, while the infrared modality is the primary modality. (c) Asymmetric cross attention module, designed to facilitate the feature interaction between the primary and auxiliary modalities.}
  \label{fig:CTFB}
\end{figure*}
\par{
  Under high-illumination conditions, pertinent information mainly centers on the visible modality, which delivers clear details and vivid colors. Conversely, under low-illumination conditions, the infrared modality becomes more crucial, capturing thermal intensity invisible to the eye. In \autoref{fig:CTFB}, specialized HI-CTFB and LI-CTFB structures help dynamically alternate between primary and auxiliary modalities and combine cross-modality information.
}
  \subsubsection{HI-CTFB and LI-CTFB}
\par{
  Initially, the aligned features $\{Z_{vi}^{in}, Z_{ir}^{in}\}\in\mathbb{R}^{H\times W\times C}$ undergo a linear projection (LP) using a shifted window strategy. This strategy divides the input into $\frac{HW}{M^2}$ non-overlapping local windows, each of size $M\times M\times C$, converting the features into a sequence of embeddings with dimensions $\frac{HW}{M^2}\times M^2\times C$. These embeddings are then forwarded to an asymmetric cross attention module and a layernorm layer, allowing the model to capture dependencies between tokens in the sequence, regardless of their spatial distance. Subsequently, a feed-forward network (FFN) with two multi-layer perceptron (MLP) layers and a gaussian error linear unit (GeLU) activation is used to extract the necessary information from the fused tokens. Moreover, the residual connection is incorporated into the full process. Mathematically, this can be expressed as:
  \begin{flalign}
    \label{eq:1}
    &Z_{ir\rightarrow vi}^{ACA} = ACA(LN(LP(Z_{vi}^{in}, Z_{ir}^{in}))) + LP(Z_{vi}^{in}), \\
    &Z_{vi\rightarrow ir}^{ACA} = ACA(LN(LP(Z_{ir}^{in}, Z_{vi}^{in}))) + LP(Z_{ir}^{in}), \\
    &Z_{ir\rightarrow vi}^{out} = FFN(LN(Z_{vi}^{ACA})) + Z_{ir\rightarrow vi}^{ACA}, \\
    &Z_{vi\rightarrow ir}^{out} = FFN(LN(Z_{ir}^{ACA})) + Z_{vi\rightarrow ir}^{ACA}.
  \end{flalign}
  where $ACA(\cdot)$ is the asymmetric cross attention operation, taking the primary modality as the first parameter and the auxiliary modality as the second. Layer normalization (LN) is applied to stabilize the activations. The outputs $\{Z_{ir\rightarrow vi}^{out}$, $Z_{vi\rightarrow ir}^{out}\}$ are stored as the input of the successive CTFB.
  }

\subsubsection{Asymmetric Cross Attention}
  \par{
    Unlike traditional multi-head self-attention limited to single modality, asymmetric cross-attention captures interactions across distinct modalities. This mechanism treats the features of the primary modality as queries and the features of the auxiliary modality as the supplementary keys and values, actively seeking out the most relevant information. Notably, these query features are inherently compatible with the key and value features due to their dimensional alignment. In \autoref{fig:CTFB}(c), each window separately runs the asymmetric cross attention after LP and LN steps. Given two local window features $\{X_{vi},X_{ir}\}\in\mathbb{R}^{M^2\times C}$, three sets of shared learnable weight matrices $\{\textbf{W}_{vi}^Q,\textbf{W}_{ir}^Q\}$, $\{\textbf{W}_{vi}^K,\textbf{W}_{ir}^K\}$, and $\{\textbf{W}_{vi}^V,\textbf{W}_{ir}^V\}$ with the same dimension $\mathbb{R}^{C\times C}$ are employed to project them into corresponding queries $\{\textbf{Q}_{vi},\textbf{Q}_{ir}\}$, keys $\{\textbf{K}_{vi},\textbf{K}_{ir}\}$, and values $\{\textbf{V}_{vi},\textbf{V}_{ir}\}$, which can be formulated as follows:
    \begin{equation}
      \begin{split}
        \label{eq:2}
        \{\textbf{Q}_{vi}, \textbf{K}_{vi}, \textbf{V}_{vi}\} &= \{X_{vi}\textbf{W}_{vi}^Q, X_{vi}\textbf{W}_{vi}^K, X_{vi}\textbf{W}_{vi}^V\}, \\
        \{\textbf{Q}_{ir}, \textbf{K}_{ir}, \textbf{V}_{ir}\} &= \{X_{ir}\textbf{W}_{ir}^Q, X_{ir}\textbf{W}_{ir}^K, X_{ir}\textbf{W}_{ir}^V\}.
      \end{split}
    \end{equation}
    Then, the dot-product of the query vector with each key vectors in the sequence is calculated. The resulting scores are normalized using the softmax operator, producing a set of attention weights. Finally, each value vector is multiplied by its respective attention weight and the results are aggregated. This procedure is formally represented as follows:
    \begin{flalign}
      &\mathbf{K} = Concat(\mathbf{K}_{vi}, \mathbf{K}_{ir}), \mathbf{V} = Concat(\mathbf{V}_{vi}, \mathbf{V}_{ir}),\\
      &Z_{ir\rightarrow vi}^{ACA} = softmax({\mathbf{Q}_{vi}\mathbf{K}^{T}}/{\sqrt{d_{k}}}+\mathbf{B_1})\mathbf{V},\\
      &Z_{vi\rightarrow ir}^{ACA} = softmax({\mathbf{Q}_{ir}\mathbf{K}^{T}}/{\sqrt{d_{k}}}+\mathbf{B_2})\mathbf{V}.
     \end{flalign}
    where $Concat(\cdot,\cdot)$ is a concatenation operation in the channel dimension, $d_k$ is the dimension of keys, $\mathbf{B_1}$ and $\mathbf{B_2}$ are the learnable relative positional encoding.
  }

\subsection{Illumination-Sensitive Gating Network} \label{method:3}
\par
\begin{figure}[!htbp]
    \centering
    \includegraphics[width=\linewidth]{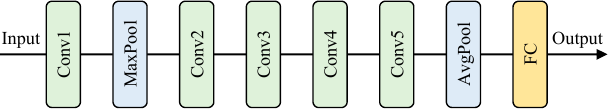}
    \caption{The structure of the illumination-sensitive gating network.}
    \label{fig:light}
  \end{figure}

{
  Following \cite{MBNet-ECCV2020}, we explore ResNet architecture \cite{resnet} for classifying the VI images into high-illumination and low-illumination categories. By carefully selecting the depth of ResNet, we aim to find a trade-off between computational efficiency and representational power. Specifically, we use ResNet18, with fewer number of residual blocks and low-dimensional convolutional layers, to accurately predict illumination distribution. As shown in \autoref{fig:light}, the network inputs visible images and outputs probability $P_H$ and $P_L$. \autoref{Table:1} also lists the weight layers.
}

\begin{table}[!htbp]
  \caption{The Detailed Structure of the ResNet18.}
  \label{Table:1}
  \resizebox{0.45\textwidth}{!}{ \begin{tabular}{ccc}
    \hline
    \textbf{Layer name} & \multicolumn{1}{c}{\textbf{Layer description}} & \multicolumn{1}{c}{\textbf{Output size}} \\ \hline
    Input        & Visible image                  & $480\times 640$    \\
    Conv1        & $[7\times7,64]$, s=2           & $240\times 320$    \\
    \multirow{3}{*}{Conv2}             & $3\times3$ Max pool, s=2                & \multirow{3}{*}{$120\times160$}    \\
            & $\left[\begin{array}{c}3\times3,64 \\3\times3,64 \\ \end{array}\right]\times2$ \\
    Conv3   & $\left[\begin{array}{c}3\times3,128 \\3\times3,128 \\ \end{array}\right]\times2$     & $60\times80$ \\ 
    Conv4   & $\left[\begin{array}{c}3\times3,256 \\3\times3,256 \\ \end{array}\right]\times2$     & $30\times40$  \\
    Conv5   & $\left[\begin{array}{c}3\times3,512 \\3\times3,512 \\ \end{array}\right]\times2$     & $15\times20$    \\
    Output       & Average Pool, Fully Connected, Sigmoid     &$1\times1$\\  \hline
  \end{tabular}}
\end{table}
\subsection{Loss Function} \label{method:4}
\subsubsection{BCE Loss for Illumination-Sensitive Sub-Network}
\par{
  The illumination-sensitive sub-network assigns the samples to HI-MoCTE or LI-MoCTE based on their illumination properties. To guide the training process, we adopt the binary cross-entropy loss function (BCE), which excels in handling binary classification problems. BCE measures the difference between the model's predicted outputs $\{P_H,P_L\}$ and the actual labels $y$. The BCE loss is defined as follows:
  \begin{equation}
    \label{bceloss}
    \mathcal{L}_{ISSN} = -ylog(P_H)-(1-y)log(P_L)
  \end{equation}
}
\subsubsection{Competitive Learning Loss for Fusion Network}
\par{
  During the training of deep learning networks for image fusion tasks, the loss functions frequently utilize intensity loss, gradient loss, and structural similarity index measure (SSIM) loss. Below are their definitions:
  \begin{flalign}
    \label{loss_eq1}
      &\mathcal{L}_{int}=\frac{1}{HW}\Vert{I_f-I_{ir}}\Vert_1 + \frac{1}{HW}\Vert{I_f-I_{vi}}\Vert_1 \nonumber \\
      &\mathcal{L}_{grad}=\frac{1}{HW}\Vert{\vert\nabla{I_f}\vert-\text{max}(\vert\nabla{I_{ir}}\vert,\vert\nabla{I_{vi}}\vert)}\Vert_1 \\
      &\mathcal{L}_{ssim}=w_1(1-ssim(I_f,I_{ir}))+w_2(1-ssim(I_f,I_{vi})) \nonumber
  \end{flalign}
  where $H$ and $W$ denote the height and width, $\Vert\cdot\Vert_1$ indicates L1-norm, $max(\cdot,\cdot)$ means element-wise maximum, $\left\lvert \cdot\right\rvert $ is absolute value operator, $\nabla$ is Sobel operator, and $ssim(\cdot,\cdot)$ refers to the structural similarity operation. $w_1$ and $w_2$ are assigned equal weights of 0.5. The total loss is formulated as follows:
  \begin{equation}
    \label{loss_eq2}
    \mathcal{L}_{total} = \alpha\mathcal{L}_{int}+ \beta\mathcal{L}_{grad}+\gamma\mathcal{L}_{ssim}
  \end{equation}
  where $\alpha$, $\beta$, and $\gamma$ serve to weigh the relative importance of sub-loss terms. However, both Eq.\eqref{loss_eq1} and Eq.\eqref{loss_eq2} ignore illumination factors. To leverage the strengths and varying performance levels of expert sub-networks, we combine the three sub-loss terms from Eq.\eqref{loss_eq1} with the illumination probability $P_i$. By dynamically adjusting the weights to experts and encouraging competition among them, each sub-loss term contributes to refining the model's understanding of illumination-related problems. It is computed as follows:
  \begin{flalign}
   & \mathcal{L}(o_i) = \mathcal{L}_{total}(I_f^i,I_{ir},I_{vi})  \label{loss_eq3}\\ 
   & \mathcal{L}_{fusion} = -\ln (\sum_{i = 1}^{N} P_ie^{-\mathcal{L}(o_i)})  \label{loss_eq4}
  \end{flalign}
  where $N$ is set to 2, representing two expert sub-network tailored to high ($P_1 = P_H$) and low ($P_2 = P_L$) illumination conditions. The partial derivative of Eq.\eqref{loss_eq4} with respect to Eq.\eqref{loss_eq3} is given as follows:
  \begin{equation}
    \frac{ \partial\mathcal{L}_{fusion}}{\partial{o_i}} = \frac{P_ie^{-\mathcal{L}(o_i)}}{ {\textstyle \sum_{j = 1}^{N}}P_je^{-\mathcal{L}(o_j)}}\mathcal{L}'(o_i) \label{loss_eq5}
  \end{equation}
  Initially, the gating network assumes equal expertise but adjusts its weights based on the predicted illumination and the performance levels of the expert sub-networks. As training iterates, it dynamically reallocates higher weights to those experts demonstrating superior accuracy. Furthermore, this loss intensifies the competitive dynamics among experts, encouraging each to outperform its peers. A lower $\mathcal{L}(o_i)$ grants a greater contribution to minimizing the overall loss. Therefore, this weighted loss function fosters a collaborative yet competitive training for handling illumination interferences.
  }
\section{Experimental Results and Analysis}\label{section4}
\subsection{Datasets}
\par{
  The DroneVehicle \cite{sun2020drone} dataset provides a comprehensive collection of multimodal aerial images captured by drones, covering diverse conditions including dark night, night, and daytime scenarios. However, due to slight misalignments in many VI and IR image pairs within this dataset, we filtered out aligned pairs based on the bounding box annotations. Thus, we constructed a refined dataset of 4,641 well-aligned image pairs for training and 256 aligned pairs for testing. The MSRS\footnote{\url{https://github.com/Linfeng-Tang/MSRS}} dataset comprises 1,444 pre-registered VI-IR image pairs, of which 1,083 pairs are used for training, while the remaining 361 pairs are reserved for evaluation. The TNO\footnote{\url{https://figshare.com/articles/dataset/TNO_Image_Fusion_Dataset/1008029}} dataset is a military-relevant collection containing 21 pre-registered VI-IR image pairs for testing. Additionally, the RoadScene\footnote{\url{https://github.com/hanna-xu/RoadScene}} dataset offers 221 pairs of pre-registered VI-IR images for testing. 
}
\subsection{Implementation Details}
\par{
  The implementation of our method runs on a machine equipped with the NVIDIA GeForce RTX 4090 GPU and the PyTorch platform. We set the window size to $M=8$ and assign the loss balance factors as $\alpha=1$, $\beta=5$, and $\gamma=10$. During training, input image pairs are resized to 128$\times$128 pixels and normalized to [0, 1]. Conversely, during testing, image pairs are permitted to keep their original resolutions. We employ the Adam optimizer with a warmup cosine annealing learning rate schedule as the optimization strategy. The model undergoes 60 epochs of training with a batch size of 8. A summary of the training procedure of our proposed method is provided in \autoref{alg:1} and \autoref{alg:2}.
  }
\begin{algorithm}[!htbp]
  \DontPrintSemicolon
  \tcc{Training the illumination-sensitive model}
  \caption{$\hat{\theta}\leftarrow$ ResNet18($I_{vi}^{1:M_{1}},y^{1:M_{1}},\theta$)}
  \label{alg:1}
  \KwIn{$\{(I_{vi}^n, y^n)\}_{n=1}^{M_1}$, a dataset of visible images.} 
  \KwIn{$\theta$, initial parameters.}
  \KwOut{$\hat{\theta}$, the trained parameters.}
  $N_{epochs}\leftarrow 100$, $\eta_1 \leftarrow (0, \infty)$\;
  \For{$i=1,2, \cdots, N_{epochs}$} 
  { 
    \For{$n=1,2, \cdots, M_1$}
    {
      $P_H,P_L\leftarrow$ ResNet18($I_{vi}^n, y^n\mid \theta$)\;
      $\mathcal{L}_{ISSN}(\theta)\leftarrow$ Eq. \eqref{bceloss}\;
      $\theta \leftarrow \theta-\eta_1 \cdot \nabla \mathcal{L}_{ISSN}(\theta)$\;
    }
  }
  \Return{$\hat{\theta}=\theta$}
\end{algorithm}
\begin{algorithm}[!htbp]
  \DontPrintSemicolon
  \tcc{Training the progressive fusion model}
	\caption{$\hat{\delta}\leftarrow$MoCTEFuse($I_{vi}^{1:M_2},I_{ir}^{1:M_2},\hat{\theta},\delta$)}
  \label{alg:2}
	\KwIn{$\{(I_{vi}^n, I_{ir}^n)\}_{n=1}^{M_2}$, a dataset of registered VI-IR image pairs.}
  \KwIn{$\hat{\theta}$, the pre-trained parameters of the illumination-sensitive sub-network.}
	\KwIn{$\delta$, initial parameters.}
  \KwOut{$\hat{\delta}$, the trained parameters.}
  ResNet18$(\cdot) \leftarrow \hat{\theta}$ and freeze\;
  $N_{epochs}\leftarrow 100$, $\eta_2 \leftarrow (0, \infty)$\;
	\For{$m=1,2, \cdots, N_{epochs}$}
	{ 
    \For{$n=1,2, \cdots, M_2$}
    {
      $P_H,P_L\leftarrow$ ResNet18($I_{vi}^n$) \;
      $I_f, I_f^H, I_f^L\leftarrow$ MoCTEFuse$(I_{vi}^n,I_{ir}^n\mid\delta)$\;

      $\mathcal{L}_{fusion}(\delta) \leftarrow$ Eq. \eqref{loss_eq4}\;
      $\delta\leftarrow \delta - \eta_2 \cdot \nabla \mathcal{L}_{fusion}(\delta)$\;
    }
	}
  \Return{$\hat{\delta}=\delta$}
\end{algorithm}
\subsection{Comparison Methods and Evaluation Metrics} \label{methods}
  \subsubsection{Comparison Methods} We conducted comprehensive comparisons with twelve state-of-the-art methods, including SwinFusion \cite{2022SwinFusion}, YDTR \cite{tang2022ydtr}, TarDAL \cite{2022TarDAL}, PIAFusion \cite{tang2022piafusion}, U2Fusion \cite{xu2020u2fusion}, DCTNet \cite{Li_DCTNet}, DATFuse \cite{tang2023datfuse}, CrossFuse \cite{Wang_CrossFuse}, SegMiF \cite{liu2023segmif}, LRRNet \cite{li2023lrrnet}, DDFM \cite{2023DDFM}, and SHIP \cite{2024ship}. Notably, all of these methods possess publicly available implementations, enabling a fair and reproducible evaluation.

  \subsubsection{Evaluation Metrics}
  We choose reference-free and reference-based metrics to assess performance. For reference-free metrics, entropy (EN) provides insights into the inherent information content of the fused images, while the standard deviation (SD) reveals the variability in contrast. For reference-based metrics, mutual information (MI) and visual information fidelity (VIF) are employed to assess the degree of information overlap and the visual fidelity between the fused image and the input images, respectively. Higher values on these indices signify better fusion performance.

  \subsection{Ablation Studies}\label{experiment2}
  To investigate the contribution of each component in the proposed MoCTEFuse on the overall performance, we conduct a series of ablation studies. Specifically, we remove or modify one component at a time while keeping the other parts unchanged, and then evaluate the model's performance on the DroneVehicle and MSRS datasets.
  
  \noindent \textbf{Effect of HI-MoCTE (w/o HI-MoCTE).} We carry out an ablation study on the HI-MoCTE module, which is pivotal in integrating and reconstructing cross-modality features under high-illumination conditions. To make sure the model works properly, we define the output of the illumination-sensitive gating network to be 0 (\textit{i.e.}, $P_H=0, P_L=1$). In \autoref{table:Ablation1}, the absence of HI-MoCTE leads to a significant drop in all four metrics, indicating its importance in our model.

  \noindent \textbf{Effect of LI-MoCTE (w/o LI-MoCTE).} We further examine the effect of the LI-MoCTE module, which is responsible for integrating and reconstructing cross-modality features under low-illumination conditions. By removing the LI-MoCTE, we can assess its contribution to the overall performance (\textit{i.e.}, $P_H=1, P_L=0$). \autoref{table:Ablation1} demonstrate that LI-MoCTE plays a vital role in enhancing the model's capabilities, as its removal leads to a decrease in performance.  

  \noindent \textbf{Effect of Competitive Loss (w/o Competitive Loss).} We also perform a ablation experiment on the competitive learning loss function in our model by replacing competitive loss function with a simpler alternative Eq.\eqref{loss_eq2}. \autoref{table:Ablation1} presents its critical contribution on the model's performance.

   \begin{table}[!htbp]
    \caption{Ablation Results of Different Components in the MoCTEFuse on the MSRS and DroneVehicle Datasets. \textbf{Bold} and \underline{Underline} Indicate the Best and Second-Best Values, Respectively.}
    \label{table:Ablation1}
    \centering
    \renewcommand{\arraystretch}{1.2} 
    \resizebox{\linewidth}{!}{
    \begin{tabular}{ccccc}
      \toprule 
      \multicolumn{5}{c}{\textbf{MSRS Dataset}}   \\
      Configurations              & EN               & SD                  & MI                & VIF \\ \midrule
      w/o HI-MoCTE   &$6.6065\pm0.7845$	&$42.4878\pm13.6999$	&$\underline{3.4835}\pm0.6616$	&$\underline{1.0196}\pm0.0969$\\                     
      w/o LI-MoCTE   &$\underline{6.6235}\pm0.7445$	&$42.0666\pm13.3137$	&$3.3977\pm0.6102$	&$0.9938\pm0.0925$\\                   
      w/o Competitive Loss        &$6.4803\pm0.9252$	&$\underline{42.4879}\pm14.6672$	&$3.2736\pm0.9703$	&$0.9369\pm0.0764$\\                      
      MoCTEFuse      &$\mathbf{6.7270}\pm0.6934$	&$\mathbf{43.1572}\pm13.3259$	&$\mathbf{3.6316}\pm0.6553$	&$\mathbf{1.0398}\pm0.1001$\\ \bottomrule          
      \multicolumn{5}{c}{\textbf{DroneVehicle Dataset}}\\
      Configurations              & EN               & SD                  & MI                & VIF \\ \midrule
      w/o HI-MoCTE              &$\underline{7.3194}\pm0.3038$	&$\underline{46.2482}\pm9.6670$	&$3.6895\pm0.7540$	&$\underline{0.8329}\pm0.1707$\\
      w/o LI-MoCTE              &$7.3085\pm0.3046$	&$45.7927\pm9.2644$	&$3.6400\pm0.6279$	&$0.8267\pm0.1686$\\
      w/o Competitive Loss      &$7.3111\pm0.2918$	&$45.7862\pm9.1345$	&$\underline{3.7582}\pm0.6940$	&$0.8279\pm0.1727$\\
      MoCTEFuse                 &$\mathbf{7.3718}\pm0.2853$	&$\mathbf{48.4722}\pm9.5070$	&$\mathbf{3.7781}\pm0.6231$	&$\mathbf{0.8550}\pm0.1638$\\  \bottomrule
      \end{tabular}
      }
    \end{table}

  \par
\begin{figure*}[!htbp]
    \centering
    \subfloat[A high illumination example (00559D) from the MSRS dataset.]{%
      \includegraphics[width=0.49\linewidth]{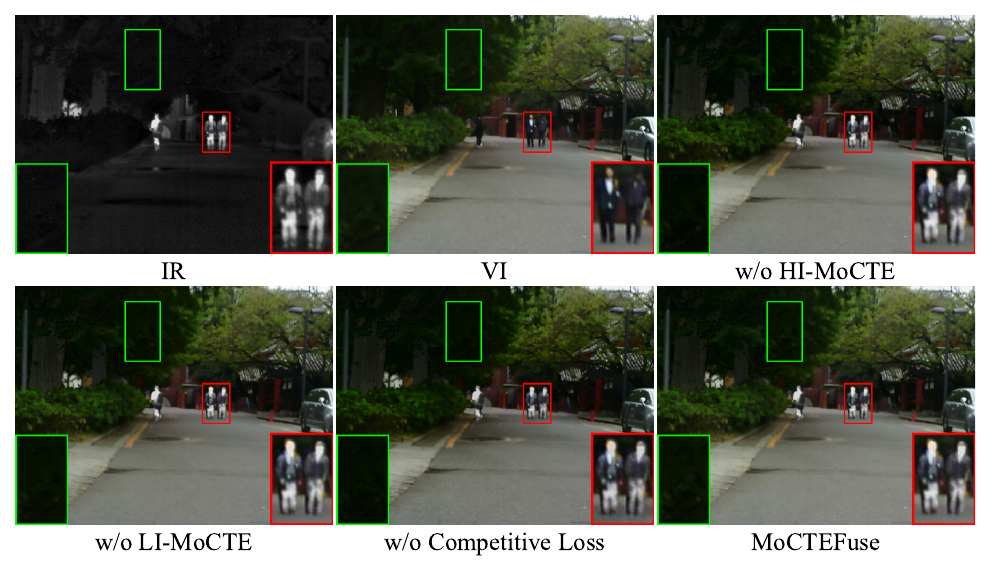}
      \label{fig:ablation1}%
      } 
    \subfloat[A low illumination example (00004N) from the MSRS dataset.]{%
      \includegraphics[width=0.49\linewidth]{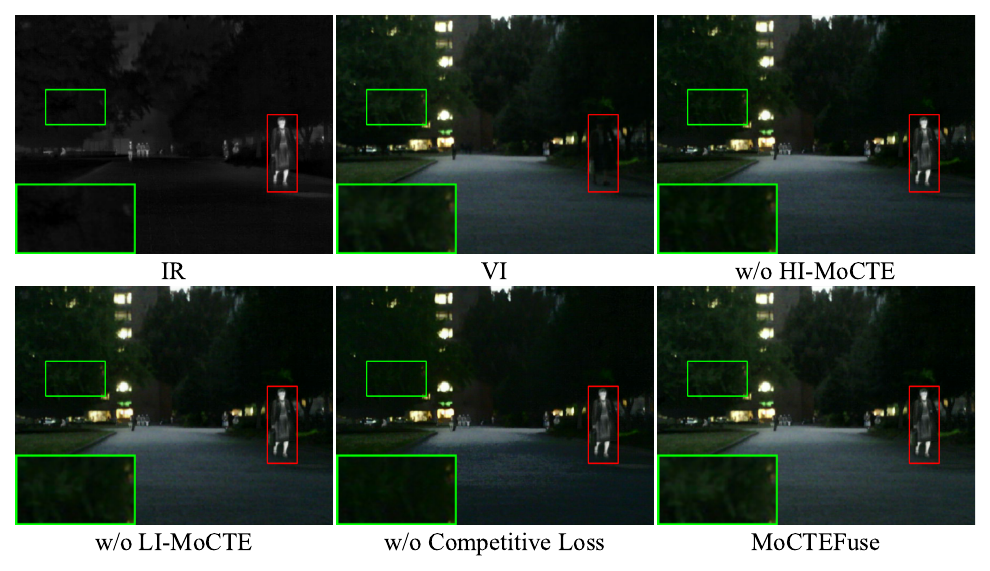}
      \label{fig:ablation2}%
      }   
    \hfill
    \subfloat[A high illumination example (val\_00833) from the DroneVehicle dataset.]{%
      \includegraphics[width=0.49\linewidth]{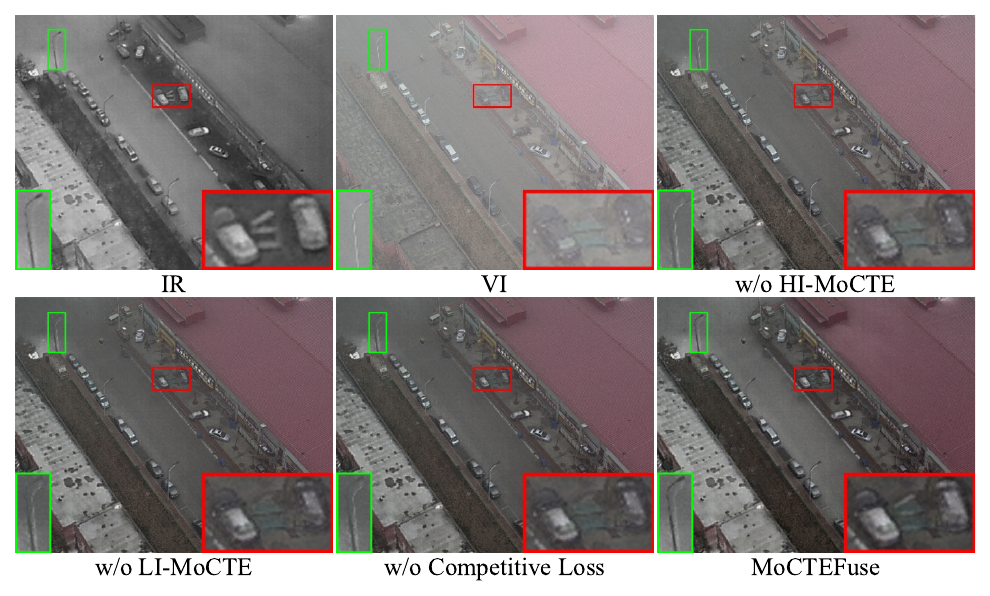}
      \label{fig:ablation3}%
      }
    \subfloat[A low illumination example (val\_01020) from the DroneVehicle dataset.]{%
      \includegraphics[width=0.49\linewidth]{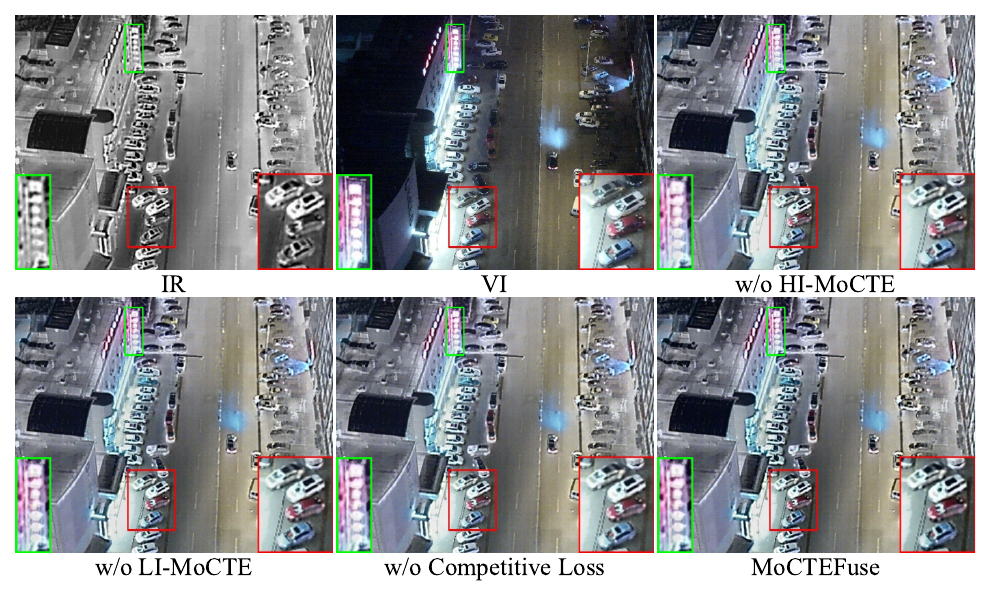}
      \label{fig:ablation4}%
      }
    \caption{Visualization of four image pairs and their fusion results under different settings.}
    \label{fig:ablation}
  \end{figure*}

{
    \autoref{fig:ablation} illustrates four pairs of source images and their respective fusion images under different configurations. These qualitative observations reinforce the quantitative performance metrics. In \autoref{fig:ablation}\subref{fig:ablation1} and \autoref{fig:ablation}\subref{fig:ablation2}, the asymmetric cross-attention module, through its primary-auxiliary mechanism for IR and VI modalities, ensures that salient targets (\textit{i.e.}, red-framed regions) remain largely unaffected across all experimental settings. However, texture details (\textit{i.e.}, green-framed regions) are subject to varying degrees of degradation, especially in the absence of the competitive loss function. In the more complex aerial scenes depicted in \autoref{fig:ablation}\subref{fig:ablation3} and \autoref{fig:ablation}\subref{fig:ablation4}, the impact of removing key components becomes more apparent, as the model struggles to effectively manage light changes. These findings reveal that the integration of HI-MoCTE, LI-MoCTE, and the competitive learning loss function enables the network to dynamically handle samples with varying illumination levels. This adaptability facilitates the learning of more representative feature mappings, leading to improved fusion performance.
  }
 
\subsection{Results on the MSRS Dataset}
\subsubsection{Qualitative Comparison}
  \par
\begin{figure*}[!htbp]
  \centering
  \subfloat[A high illumination example (00630D)]{%
    \includegraphics[width=0.49\linewidth]{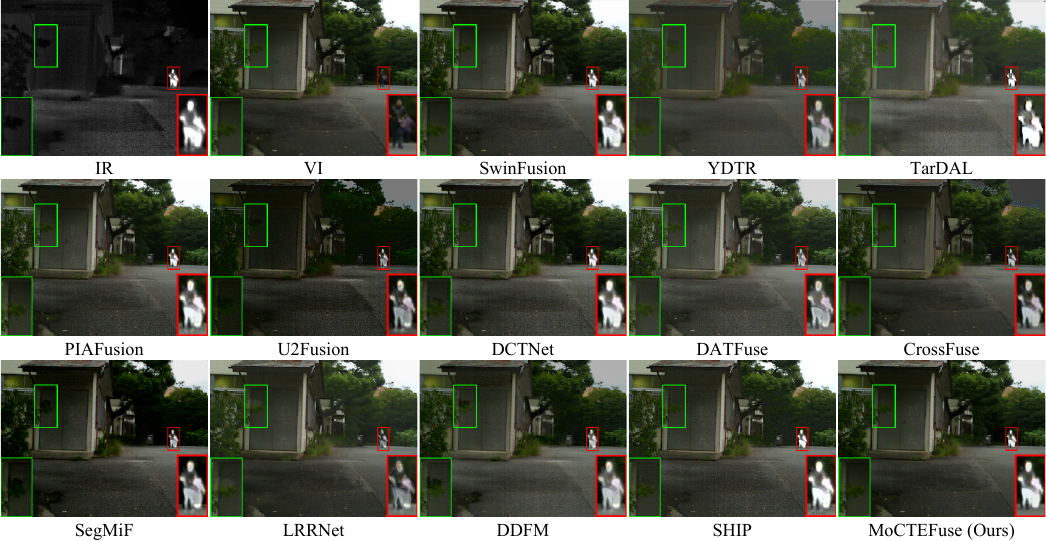}
    \label{fig:00630D}%
    } \hfill
  \subfloat[A low illumination example (00730N)]{%
    \includegraphics[width=0.49\linewidth]{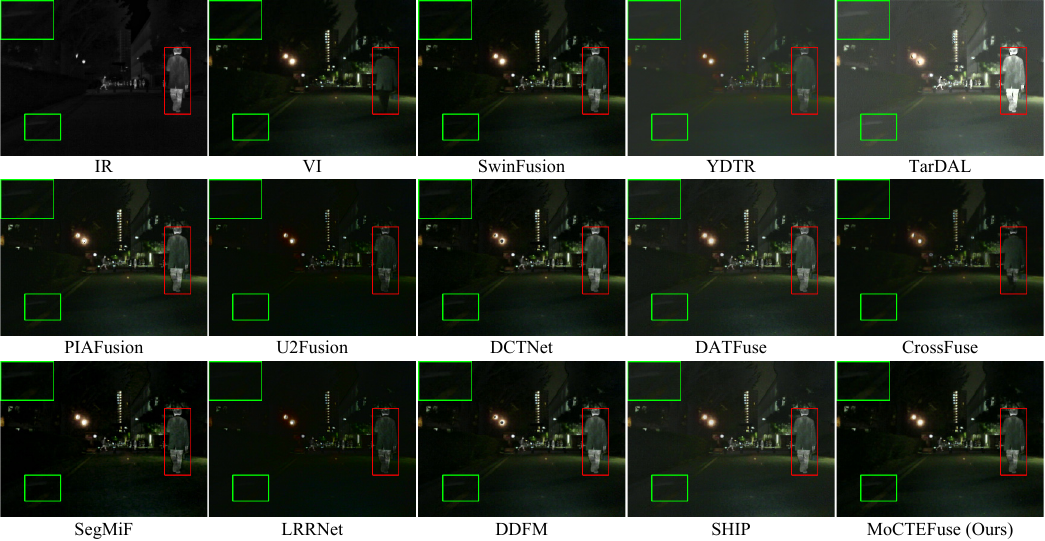}
    \label{fig:00730N}%
    }
  \caption{Visual comparisons of IR, VI, and fused images produced by twelve state-of-the-art methods with MoCTEFuse on the MSRS dataset.}
  \label{fig:MSRS}
\end{figure*}

{\autoref{fig:MSRS} depicts two typical examples of fused images. Upon close examination, these images reveal distinct characteristics and differences in the fusion process for each method. To improve visualization, enlarged views of texture details and salient objects within the green- and red-framed regions are provided.}

  \par{In the high-illumination scene of \autoref{fig:MSRS}\subref{fig:00630D}, the VI image is rich in texture details, while the IR image exhibits abundant salient target information, albeit with accompanying some noise. To achieve optimal visual quality, fusing these images requires preserving nuanced textural details from the VI image, incorporating salient target information from the IR image, and minimizing noise. U2Fusion produces a fusion image with an unwished dark background that lacks image details. YDTR and TarDAL clearly present the salient targets (\textit{i.e.}, red-framed regions) from the source images. However, there are some minor deficiencies in visual fidelity, leading to unnatural and blurred fused outputs. CrossFuse introduces undesirable distortions around trees. LRRNet and DDFM exhibit artifacts in the green-framed regions. In contrast, SwinFusion, DCTNet, DATFuse, SegMiF, and SHIP perform well in terms of visual fidelity and image details, successfully retaining a lot of meaningful information from the source images. However, they are coarse in handling image contrast, introducing excessive noise interference. Compared to aforementioned competitors, MoCTEFuse yields a visually appealing images, effectively addressing the balanced preservation of texture details and salient targets, while reducing noise.}
\par{
  In the low-illumination scene of \autoref{fig:MSRS}\subref{fig:00730N}, the quality of VI image is obviously compromised in its ability to render attractive texture details, whereas the IR image maintains its advantage in capturing salient targets. However, the negative effects of U2Fusion, YDTR, and TarDAL in the high-illumination scene persist, diminishing the visual quality of the fused images. Compared to SwinFusion, PIAFusion, DCTNet, DATFuse, CrossFuse, SegMiF, LRRNet, DDFM, and SHIP, our fused images show better texture details (\textit{i.e.}, green-framed regions). In addition, our fused images display satisfactory salient targets. Notably, TarDAL produces a brighter appearance, particularly in the depiction of the people (\textit{i.e.}, red-framed regions). Although this gray enhancement may initially appear appealing, it readily causes a decrease in contrast under high-illumination conditions.
}

\begin{table*}[!htbp]
  \caption{Quantitative Comparisons of MoCTEFuse with 12 State-of-the-Arts on the Four Datasets. The Best Results are Marked in \textbf{Bold}.}
  \label{table:MSRS}
  \centering
  \renewcommand{\arraystretch}{1.2} 
  \resizebox{0.95\linewidth}{!}{
  \begin{tabular}{cc|cccc|cccc}
  \toprule
  \multirow{2}{*}{Methods} & \multirow{2}{*}{Venues} & \multicolumn{4}{c|}{\textbf{MSRS Dataset}}   & \multicolumn{4}{c}{\textbf{DroneVehicle Dataset}} \\
                                     &             & EN               & SD                  & MI                & VIF 
                                                   & EN               & SD                  & MI                & VIF \\ \cmidrule{1-10}
  SwinFusion \cite{2022SwinFusion}	 &JAS'2022     &$6.6219\pm0.7825$	&$42.8339\pm13.1527$	&$3.1594\pm0.8554$	&$0.9884\pm0.0669$
                                                   &$7.3516\pm0.3192$	&$48.2938\pm10.5829$	&$2.0773\pm0.3315$	&$0.6232\pm0.0892$\\
  YDTR \cite{tang2022ydtr}	         &TMM'2022     &$5.6451\pm1.0531$	&$25.3704\pm12.1359$	&$1.9181\pm0.5453$	&$0.5781\pm0.0941$
                                                   &$7.0682\pm0.3267$	&$40.0001\pm8.4104$	  &$1.9363\pm0.3335$	&$0.5078\pm0.0645$ \\
  TarDAL \cite{2022TarDAL}	         &CVPR'2022    &$6.4770\pm0.6300$	&$37.5936\pm8.7087$	  &$1.8248\pm0.4366$	&$0.6948\pm0.1199$
                                                   &$6.7180\pm1.0253$	&$45.3847\pm13.8225$	&$1.6546\pm0.4390$	&$0.4198\pm0.0993$ \\
  PIAFusion \cite{tang2022piafusion} &INFFUS'2022  &$6.5710\pm0.8440$	&$42.5371\pm13.3480$	&$3.1719\pm0.8146$	&$0.9877\pm0.0782$
                                                   &$7.2819\pm0.3624$	&$44.6451\pm10.1695$	&$1.9144\pm0.3998$	&$0.5262\pm0.0892$ \\
  U2Fusion \cite{xu2020u2fusion}	   &TPAMI'2020   &$5.3742\pm1.1735$	&$25.4950\pm9.7765$	  &$1.4003\pm0.3022$	&$0.5418\pm0.0690$
                                                   &$6.9153\pm0.4302$	&$36.3894\pm9.7334$	  &$1.6032\pm0.2704$	&$0.4516\pm0.0496$ \\
  DCTNet \cite{Li_DCTNet}	           &TIM'2023     &$6.7032\pm0.7392$	&$42.6050\pm13.2225$	&$2.9428\pm0.8882$	&$1.0349\pm0.1145$
                                                   &$7.3226\pm0.3260$	&$46.3492\pm9.9866$	  &$1.9221\pm0.4353$	&$0.5334\pm0.0910$ \\
  DATFuse \cite{tang2023datfuse} 	   &TCSVT'2023   &$6.4796\pm0.7335$	&$36.4757\pm10.5737$	&$2.6963\pm0.5769$	&$0.9063\pm0.1077$
                                                   &$7.0737\pm0.3626$	&$42.1847\pm9.6956$	  &$1.7977\pm0.2566$	&$0.402\pm0.0544$ \\
  CrossFuse \cite{Wang_CrossFuse}	   &TCSVT'2023   &$5.7578\pm0.8235$	&$23.3004\pm5.8253$	  &$2.7332\pm0.4998$	&$0.6181\pm0.0964$
                                                   &$6.9717\pm0.3574$	&$38.7255\pm8.8742$	  &$2.0703\pm0.2701$	&$0.5113\pm0.0596$ \\
  SegMiF \cite{liu2023segmif}        &ICCV'2023    &$6.4015\pm0.9836$	&$41.9585\pm14.6953$	&$1.7800\pm0.4577$	&$0.7575\pm0.0927$
                                                   &$7.1829\pm0.4448$	&$45.4817\pm12.9053$	&$1.7308\pm0.3778$	&$0.5068\pm0.0735$ \\
  LRRNet \cite{li2023lrrnet}         &TPAMI'2023   &$6.1919\pm0.8694$	&$31.7580\pm11.4316$	&$2.0255\pm0.5279$	&$0.5412\pm0.0957$
                                                   &$6.6224\pm0.4258$	&$29.1323\pm7.8308$	  &$1.4183\pm0.3090$	&$0.4134\pm0.0998$ \\
  DDFM    \cite{2023DDFM}            &ICCV'2023    &$6.1747\pm0.7512$	&$28.9176\pm8.6541$	  &$1.8914\pm0.3398$	&$0.7429\pm0.1163$
                                                   &$6.9573\pm0.3779$	&$35.9265\pm9.1319$	  &$1.8430\pm0.3937$	&$0.5470\pm0.0985$ \\
  SHIP    \cite{2024ship}            &CVPR'2024    &$6.4308\pm0.8995$	&$41.1310\pm13.2439$	&$2.8613\pm1.0294$	&$0.9075\pm0.0739$
                                                   &$7.3045\pm0.3397$	&$45.5152\pm9.9038$	  &$2.2866\pm0.3623$	&$0.6471\pm0.1124$  \\
  MoCTEFuse (Ours)       	           &-            &$\mathbf{6.7270}\pm0.6934$	&$\mathbf{43.1572}\pm13.3259$	&$\mathbf{3.6316}\pm0.6553$	&$\mathbf{1.0398}\pm0.1001$
                                                   &$\mathbf{7.3718}\pm0.2853$	&$\mathbf{48.4722}\pm9.5070$	&$\mathbf{3.7781}\pm0.6231$	&$\mathbf{0.8550}\pm0.1638$ \\
  \midrule
  \multirow{2}{*}{Methods} & \multirow{2}{*}{Venues} & \multicolumn{4}{c|}{\textbf{TNO Dataset}}   & \multicolumn{4}{c}{\textbf{RoadScene Dataset}} \\
                                     &             & EN               & SD                  & MI                & VIF 
                                                   & EN               & SD                  & MI                & VIF \\ \cmidrule{1-10}
  SwinFusion \cite{2022SwinFusion}	 &JAS'2022     &$6.7090\pm0.4959$	&$38.0902\pm12.4528$	&$2.2692\pm0.5256$	&$0.7192\pm0.1148$
                                                   &$6.9827\pm0.3266$	&$44.1076\pm7.0436$	  &$2.3532\pm0.3705$	&$0.6252\pm0.0943$ \\
  YDTR \cite{tang2022ydtr}	         &TMM'2022     &$6.2268\pm0.4730$	&$24.0541\pm7.3115$	  &$1.6745\pm0.4633$	&$0.5461\pm0.0966$
                                                   &$6.9137\pm0.2955$	&$35.9004\pm7.4243$	  &$2.0415\pm0.4046$	&$0.5570\pm0.1004$ \\
  TarDAL \cite{2022TarDAL}	         &CVPR'2022    &$6.9135\pm0.5595$	&$43.2834\pm11.0049$	&$1.9102\pm0.4017$	&$0.6050\pm0.0953$
                                                   &$7.2953\pm0.2559$	&$46.2435\pm6.6591$	&$2.3747\pm0.2957$	&$0.5582\pm0.0862$ \\
  PIAFusion \cite{tang2022piafusion} &INFFUS'2022  &$6.7425\pm0.5174$	&$39.5406\pm13.9012$	&$2.3365\pm0.5215$	&$0.7326\pm0.1133$
                                                   &$7.1252\pm0.2006$	&$42.5977\pm6.1256$	&$2.2178\pm0.4304$	&$0.5845\pm0.1055$ \\
  U2Fusion \cite{xu2020u2fusion}	   &TPAMI'2020   &$6.7571\pm0.4111$	&$31.7083\pm7.8011$	  &$1.2451\pm0.2986$	&$0.5829\pm0.1213$
                                                   &$7.2642\pm0.2659$	&$42.8958\pm8.7958$	&$1.9314\pm0.3903$	&$0.5830\pm0.1078$ \\
  DCTNet \cite{Li_DCTNet}	           &TIM'2023     &$\mathbf{6.9618}\pm0.3764$	&$40.6216\pm12.2068$	&$1.9882\pm0.4923$	&$0.7550\pm0.1666$
                                                   &$7.2289\pm0.2031$	&$43.5599\pm6.7486$	&$1.8639\pm0.3513$	&$0.5990\pm0.1202$ \\
  DATFuse \cite{tang2023datfuse} 	   &TCSVT'2023   &$6.3206\pm0.4310$	&$26.4445\pm8.3365$	&$2.3407\pm0.4554$	&$0.6716\pm0.0965$
                                                   &$6.7200\pm0.2797$	&$31.7475\pm5.6176$	&$2.5405\pm0.3522$	&$0.5966\pm0.0875$ \\
  CrossFuse \cite{Wang_CrossFuse}	   &TCSVT'2023   &$6.6109\pm0.5826$	&$30.9257\pm8.8739$	  &$3.0791\pm0.4854$	&$0.7800\pm0.0741$
                                                   &$6.8012\pm0.2083$	&$32.4120\pm5.8240$	&$3.1565\pm0.5392$	&$0.7062\pm0.1131$ \\
  SegMiF \cite{liu2023segmif}        &ICCV'2023    &$6.8674\pm0.5633$	&$43.8695\pm12.9291$	&$2.1393\pm0.4353$	&$0.8366\pm0.1394$
                                                   &$7.3561\pm0.2415$	&$49.8691\pm8.2598$	&$1.8900\pm0.3835$	&$0.6193\pm0.0947$ \\
  LRRNet \cite{li2023lrrnet}         &TPAMI'2023   &$6.8384\pm0.5729$	&$39.4994\pm13.0721$	&$1.7436\pm0.3214$	&$0.5508\pm0.1101$
                                                   &$7.1315\pm0.2085$	&$42.4370\pm5.8933$	&$1.9408\pm0.3349$	&$0.4938\pm0.0898$ \\
  DDFM    \cite{2023DDFM}            &ICCV'2023    &$6.7302\pm0.4195$	&$32.7349\pm9.4987$	  &$1.4837\pm0.3473$	&$0.6116\pm0.1166$
                                                   &$7.1644\pm0.2153$	&$43.3941\pm6.8536$	&$2.5823\pm0.3770$	&$0.6292\pm0.1146$ \\
  SHIP    \cite{2024ship}            &CVPR'2024    &$6.7493\pm0.5336$	&$37.5745\pm11.9962$	&$2.6188\pm0.4471$	&$0.7199\pm0.1146$
                                                   &$6.4308\pm0.8995$	&$41.1310\pm13.2439$	&$2.8613\pm1.0294$	&$0.9075\pm0.0739$  \\
  MoCTEFuse (Ours)       	           &-            &$6.8572\pm0.4619$	&$\mathbf{45.0258}\pm13.1315$	&$\mathbf{3.3803}\pm0.4829$	&$\mathbf{0.9578}\pm0.2240$
                                                   &$\mathbf{7.4183}\pm0.2032$	&$\mathbf{55.9930}\pm7.5811$	&$\mathbf{3.5419}\pm0.3371$	&$\mathbf{0.8390}\pm0.1188$ \\
  \bottomrule
  \end{tabular}
  }
  \end{table*}
  \begin{figure*}
    \subfloat[Statistical results of all 361 image pairs from the MSRS dataset]{
      \centering
      \includegraphics[width=0.95\linewidth]{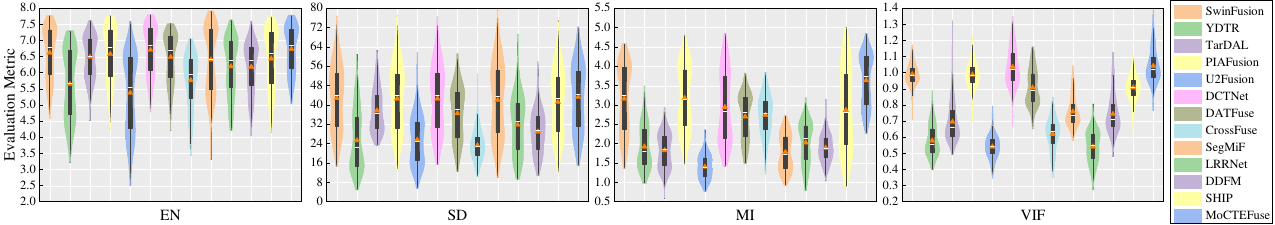}
      \label{fig:a}
      }\hfil
    \subfloat[Statistical results of all 256 image pairs from the DroneVehicle dataset]{
      \centering
      \includegraphics[width=0.95\linewidth]{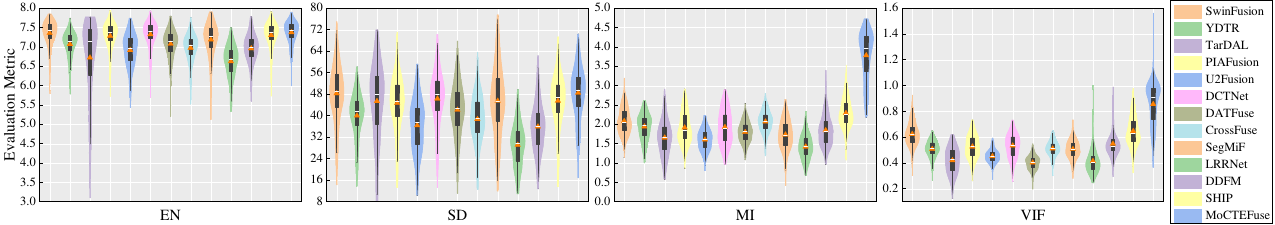}
      \label{fig:b}
      }\hfil
    \subfloat[Statistical results of all 21 image pairs from the TNO dataset]{
      \centering
      \includegraphics[width=0.95\linewidth]{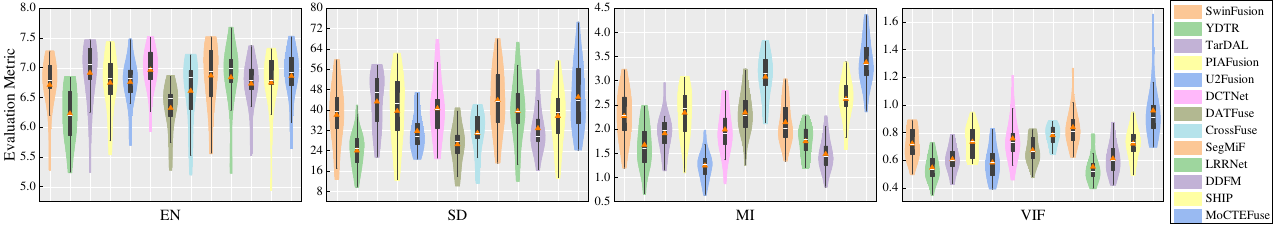}
      \label{fig:c}
      }\hfil
      \subfloat[Statistical results of all 221 image pairs from the RoadScene dataset]{
      \centering
      \includegraphics[width=0.95\linewidth]{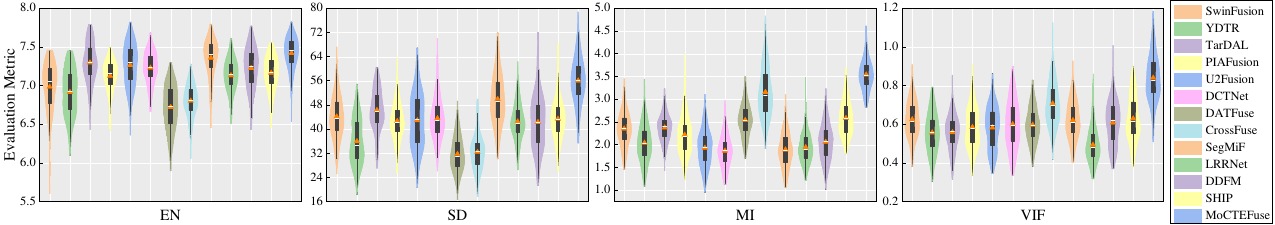}
      \label{fig:d}
      }
    \caption{Quantitative comparisons of thirteen methods on four public datasets. The x-axis represents different metrics, while the y-axis shows the corresponding values. Within the violins diagrams, the white lines denote the median, and the orange triangles mark the mean.}
    \label{fig:allimgs}
  \end{figure*}
\subsubsection{Quantitative Comparison}
  \par{
    To quantitatively evaluate the performance of our MoCTEFuse, we test it on all the 361 image pairs of the MSRS dataset. We also repeat the experiments several times and perform statistical analysis on the results to ensure the robustness and reliability of the experimental results. As can be seen from the \autoref{table:MSRS}, MoCTEFuse achieves the highest mean values in all four metrics. These superior metrics generally suggest that MoCTEFuse generates images characterized by rich information content, smooth contrast transitions, minimal redundancy in the fusion process, and a heightened alignment with human visual perception, respectively. Meanwhile, \autoref{fig:allimgs}\subref{fig:a} offers a detailed visualization of the test sample distribution and probability density for each metric. Notably, MoCTEFuse consistently gets the highest median values across four metrics. The breadth of the violin plot shows that the index values of most fused images generated by our method exceed those produced by others.
  }

\subsection{Results on the DroneVehicle Dataset}
\subsubsection{Qualitative Comparison}
\begin{figure*}[!htbp]
  \centering
  \subfloat[A high illumination example (val\_01292)]{%
    \includegraphics[width=0.49\linewidth]{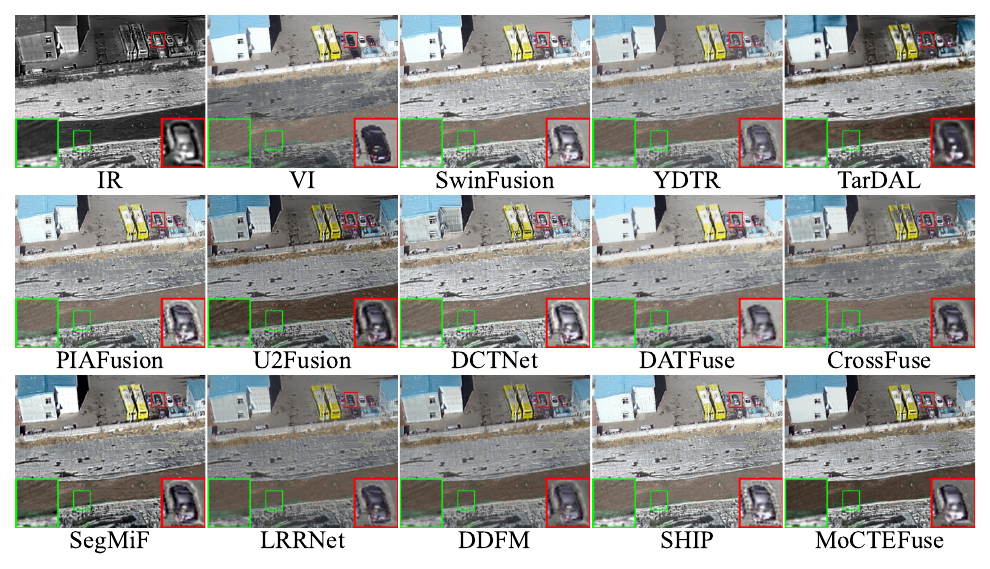}
    \label{fig:val_01292}%
    } \hfill
  \subfloat[A low illumination example (val\_01198)]{%
    \includegraphics[width=0.49\linewidth]{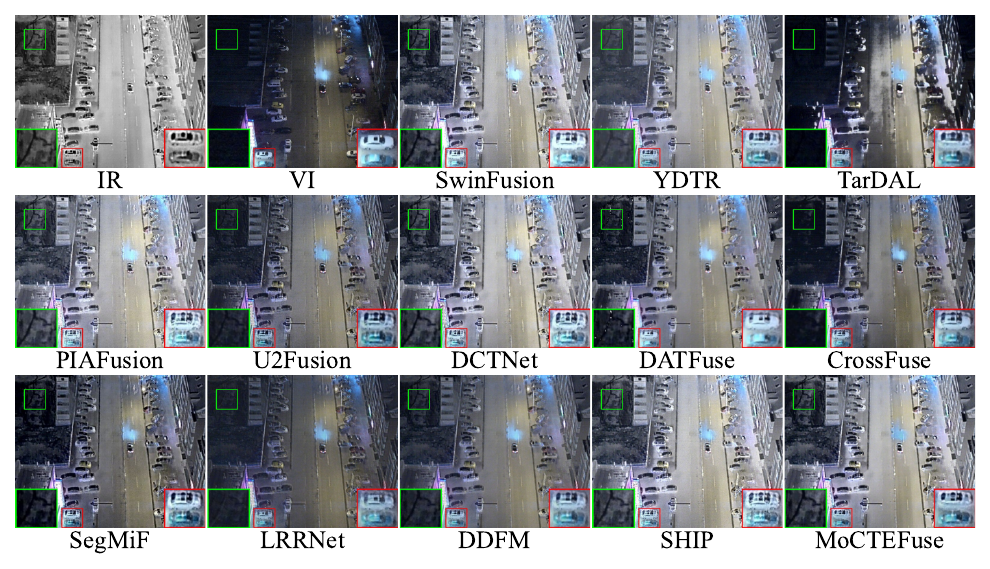}
    \label{fig:val_01198}%
    }
  \caption{Visual comparisons of IR, VI, and fused images produced by twelve state-of-the-art methods with MoCTEFuse on the DroneVehicle dataset.}
  \label{fig:DroneVehicle}
\end{figure*}

To assess the effectiveness of MoCTEFuse, we further conduct comparative experiments using various methods on the DroneVehicle dataset. \autoref{fig:DroneVehicle} displays two representative examples of fused images generated by thirteen state-of-the-art methods. To enhance clarity and facilitate intuitive analysis, we include magnified views of the texture details and salient objects within the regions highlighted in green and red. 

As illustrated in \autoref{fig:DroneVehicle}\subref{fig:val_01292}, achieving a balance between completeness and avoiding redundancy in cross-modality information remains challenging for most fusion methods. For instance, the IR image exhibits blurry artifacts around the car, which can be attributed to the bird's-eye view perspective during the imaging process. In contrast, the VI image shows accurate edge contours without any undesired artifacts. However, despite the application of various fusion methods, including SwinFusion, YDTR, TarDAL, PIAFusion, U2Fusion, DCTNet, CrossFuse, SegMiF, LRRNet, DDFM, and SHIP, redundant artifacts surrounding the car persist and remain unacceptable. While DATFuse yields clearer outlines of the car, the details and contrast are still unsatisfactory and lag behind the results produced by MoCTEFuse.

By analyzing the \autoref{fig:DroneVehicle}\subref{fig:val_01198}, we observe that other methods produce severe ghosting (\textit{i.e.}, red-framed regions), whereas our MoCTEFuse effectively avoids this issue. This distinction underscores the importance of accurately preserving detail and edge information in fusing weak misaligned image pairs. Ghosting not only affects the visual quality of the images but can also negatively impact subsequent image analysis and interpretation. In contrast, our MoCTEFuse successfully reduces redundant information during the fusion process, enhancing the clarity and recognizability of the fused images.

\subsubsection{Quantitative Comparison}
\autoref{table:MSRS} summarizes the performance metrics of thirteen image fusion methods, including EN, SD, MI, and VIF. The data clearly indicate that our MoCTEFuse outperforms all competitors across all four metrics achieving scores of 6.7270, 43.1572, 3.6316, and 1.0398. In contrast, U2Fusion demonstrates the lowest performance, with scores of 5.3742, 25.4950, 1.4003, and 0.5418. These results underscore MoCTEFuse's effectiveness in preserving the original input information and fidelity. The statistical results depicted in \autoref{fig:allimgs}\subref{fig:b} further support these findings and suggest that MoCTEFuse is a viable option for applications in aerial scenes, while other methods may need further refinement and optimization to achieve comparable performance.
 \begin{figure*}[!htbp]
  \centering
  \subfloat[A high illumination example (Tree\_sequence)]{%
    \includegraphics[width=0.49\linewidth]{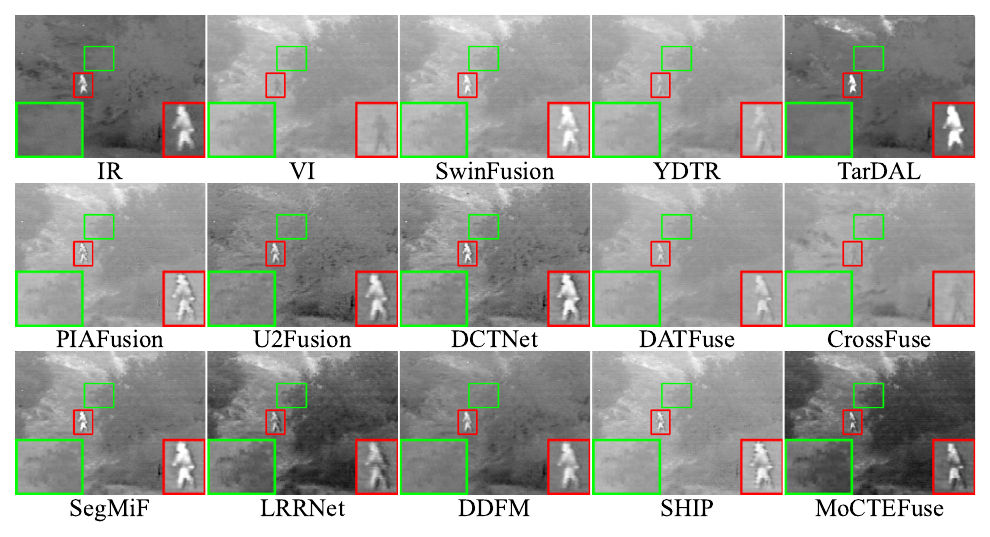}
    \label{fig:17}%
  }\hfill
  \subfloat[A low illumination example (Movie\_18)]{%
    \includegraphics[width=0.492\linewidth]{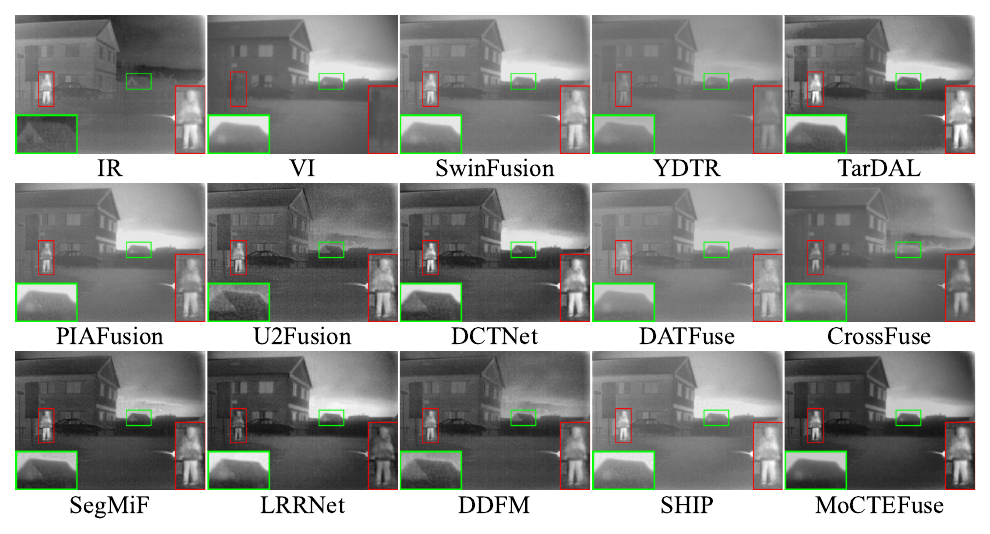} 
    \label{fig:10}%
    }
  \caption{Visual comparisons of IR, VI, and fused images produced by twelve state-of-the-art methods with MoCTEFuse on the TNO dataset.}
  \label{fig:TNO}
\end{figure*}

\subsection{Results on the TNO and RoadScene Datasets}
Generalization performance is a key aspect of evaluating deep learning methods. This section presents the generalization experiments where our model is directly applied to the TNO and RoadScene datasets after being trained on the MSRS dataset, without any additional modifications or fine-tuning.

\subsubsection{Qualitative Comparison} 
  \par{
    \autoref{fig:TNO} reports two sets of source image pairs from the TNO dataset and their fused images achieved by various methods. While these methods demonstrate good performance, there remain certain gaps compared to our proposed MoCTEFuse. For instance, YDTR and CrossFuse encounter challenges in adequately preserving valuable information from the input images, resulting in indistinct targets and blurred scenes. Undesirable noise interference occurs in U2Fusion, DCTNet, and SHIP. SwinFusion, TarDAL, PIAFusion, DATFuse, and DDFM exhibit limited capabilities in capturing intricate texture details. Although SegMiF and LRRNet alleviate the aforementioned issues to some extent, their fusion images still suffer from varying degrees of spectral pollution. In contrast, our method produces fusion images that are more informative and exhibit a significantly higher contrast, which aligns well with human visual perception. Specifically, our approach maximizes intensity information from IR images to highlight salient targets of interest. Meanwhile, it preserves rich texture details from VI images to create a clear and comprehensive backgrounds.
  }
  \begin{figure}[!htbp]
    \subfloat[A high illumination example (FLIR\_08835)]{
      \includegraphics[width=\linewidth]{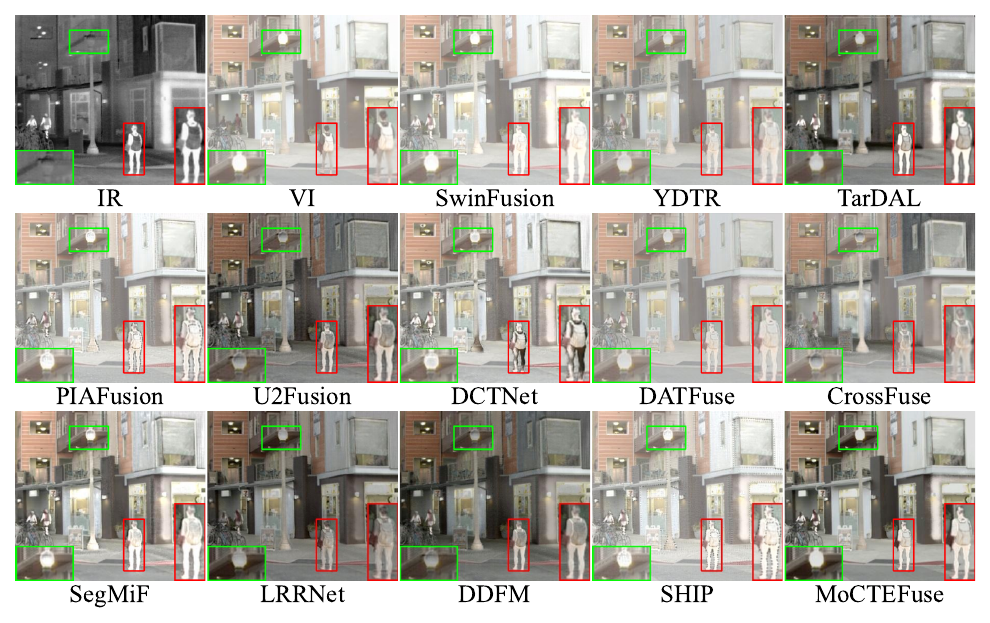}
      \label{fig:08835}
    }\vspace{-2mm}
    \subfloat[A low illumination example (FLIR\_09367)]{
      \includegraphics[width=\linewidth]{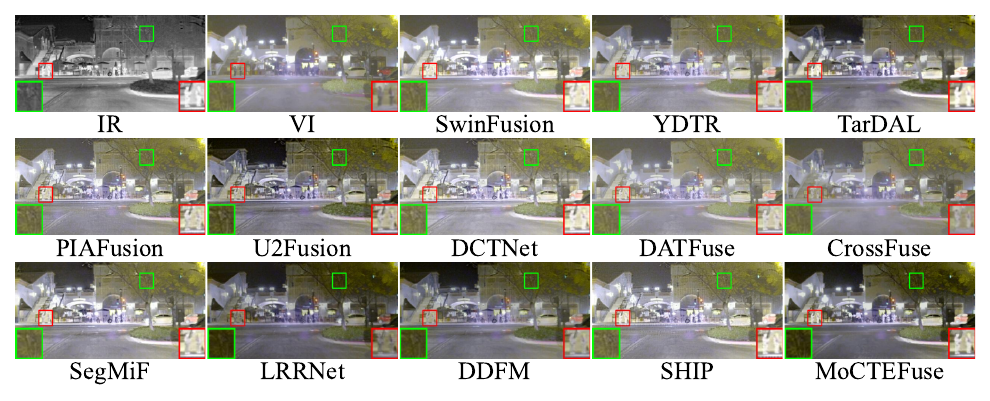}
      \label{fig:09367}
      }
    \caption{Visual comparisons of IR, VI, and fused images produced by twelve state-of-the-art methods with MoCTEFuse on the RoadScene dataset.}
    \label{fig:RoadScene}
  \end{figure}
  \par{
    In addition, two typical scenes from the RoadScene dataset and their fusion results using various methods are illustrated in \autoref{fig:RoadScene}. Upon careful evaluation, the subjective quality of the fused images obtained by our MoCTEFuse surpasses that of other alternatives. Firstly, the overall visual appearance is clear, with suitable brightness and contrast, free from blurring or distortion. Secondly, edge and texture information are abundant, accurately reflecting the details and features of the source images. Furthermore, the colors appear natural, and salient targets are effectively emphasized, with no introduction of redundant noise, ensuring the reliability of the images.
  }
  \begin{figure*}[!htbp]
    \centering
    \includegraphics[width=0.62\linewidth]{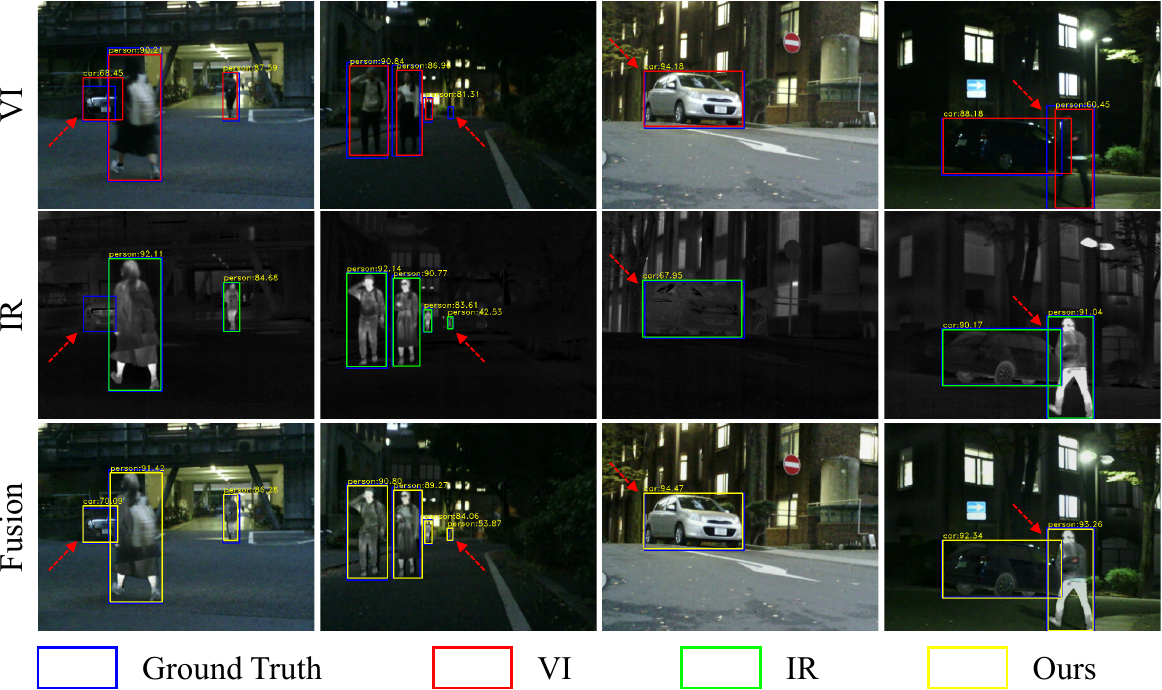}
    \caption{Visual results of general object detection. From left to right are four representative scenes. From top to bottom are the detection results of visible images, infrared images and fused images of our MoCTEFuse.}
    \label{fig:mfnet}
  \end{figure*}
  \begin{figure*}[!htbp]
    \subfloat[A high illumination example (val\_00968)]{%
      \centering
      \setlength{\abovecaptionskip}{0.cm}
      \includegraphics[width=\linewidth]{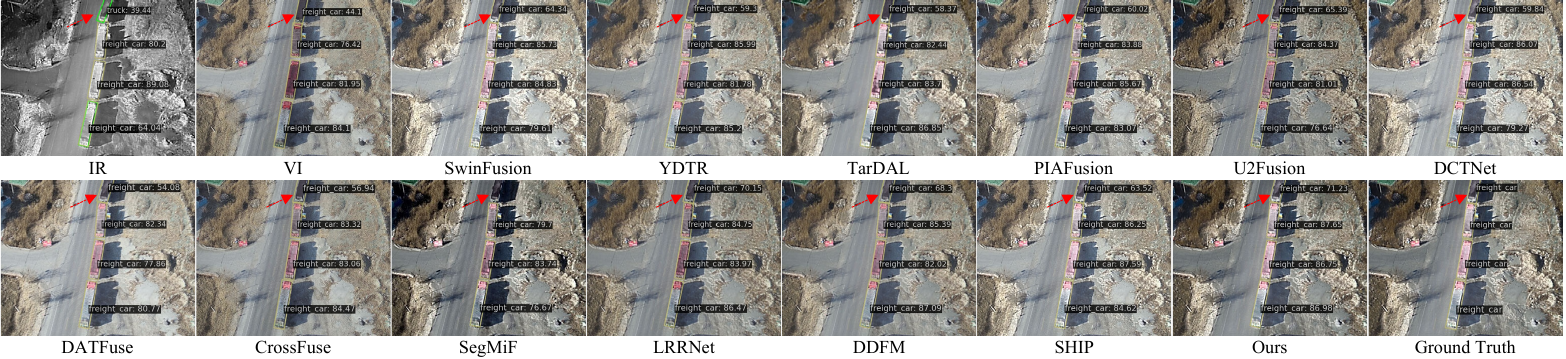}
      \label{fig:droneD}%
      }\vspace{-2mm}
    \subfloat[A low illumination example (val\_01468)]{%
      \centering
      \setlength{\abovecaptionskip}{0.cm}
      \includegraphics[width=\linewidth]{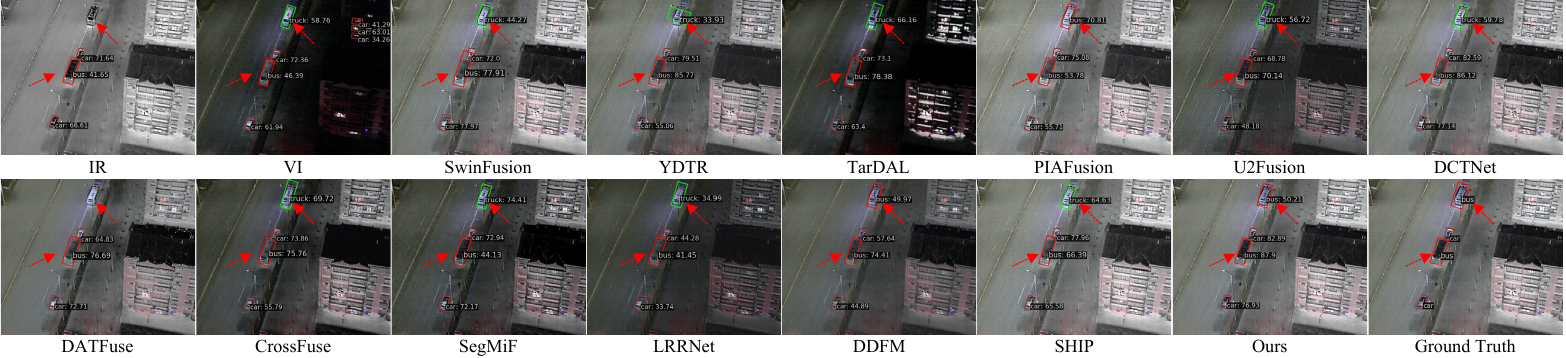}
      \label{fig:droneN}%
      }
    \caption{Visual results of rotating target detection for the DroneVehicle dataset.}
    \label{fig:drones}
  \end{figure*}
\begin{table*}[!htbp]
    \caption{Class-wise AP of VI, IR, and Fused Images Obtained From Thirteen State-of-the-Arts on the MFNet and DroneVehicle Datasets. The Best and Second-Best Results are Marked in \textbf{Blod} and \underline{Underline}.}
    \label{table:detection}
    \centering
    \renewcommand{\arraystretch}{1.2} 
    \resizebox{\linewidth}{!}{
    \begin{tabular}{cc|cccccc|cccccccccccc}
    \toprule
    \multirow{2}{*}{Methods} & \multirow{2}{*}{Venues} & \multicolumn{6}{c|}{\textbf{MFNet Dataset}}   & \multicolumn{12}{c}{\textbf{DroneVehicle Dataset}} \\ \cmidrule{3-20}
                                       &             & person    & car    & \multicolumn{1}{c|}{AP$_{0.5}$}  
                                                     & person    & car    & AP$_{0.5:0.95}$
                                                     & car	&freight car	&truck	&bus	&van   & \multicolumn{1}{c|}{AP$_{0.5}$}
                                                     & car	&freight car	&truck	&bus	&van   & AP$_{0.5:0.95}$   \\ \cmidrule{1-20}
      VI                      &-           &0.6760 &0.8622 &\multicolumn{1}{c|}{0.7691} &0.3601 &0.6824 &0.5212
                                           &0.5275 &0.2745 &0.2740 &0.5213 &0.2358 &\multicolumn{1}{c|}{0.3666} 	&0.3099 	&0.1638 	&0.1577 	&0.3366 	&0.1500 	&0.2236\\
      IR                      &-           &\textbf{0.9381} &0.6720 &\multicolumn{1}{c|}{0.8051} &\textbf{0.7031} &0.5401 &0.6216
                                           &0.8816 	&0.3475 	&0.3560 	&0.7988 	&0.3385 	&\multicolumn{1}{c|}{0.5445} 	&0.5187 	&0.1579 	&0.1497 	&0.5203 	&0.2193 	&0.3132\\
      SwinFusion \cite{2022SwinFusion}	 &JAS'2022  &0.8966 &0.8464 &\multicolumn{1}{c|}{0.8715} &0.6111 &0.6915 &0.6513
                                           &0.8943 	&0.6125 &0.4820 &0.8985 	&\underline{0.5870} 	&\multicolumn{1}{c|}{0.6949} 	&0.5724 	&0.3202 	&0.2576 	&0.6383 	&0.3637 	&0.4305\\
      YDTR \cite{tang2022ydtr}	         &TMM'2022  &0.9070 &0.8645 &\multicolumn{1}{c|}{0.8858} &0.5968 &0.6889 &0.6428
                                           &0.8926 	&0.6336 &0.4909 &0.8949 &0.5234 &\multicolumn{1}{c|}{0.6871} 	&0.5685 	&\underline{0.3444} 	&0.2637 	&0.6389 	&0.3404 	&0.4312\\
      TarDAL \cite{2022TarDAL}	         &CVPR'2022   &0.8780 &0.8313 &\multicolumn{1}{c|}{0.8547} &0.6121 &0.6590 &0.6356
      &0.6192 	&0.2902 	&0.2951 	&0.6085 	&0.2804 	&\multicolumn{1}{c|}{0.4187} 	&0.3785 	&0.1748 	&0.1849 	&0.3903 	&0.1707 	&0.2598\\
      PIAFusion \cite{tang2022piafusion} &INFFUS'2022 &\underline{0.9376} &0.8665 &\multicolumn{1}{c|}{0.9021} &\underline{0.6763} &0.7087 &0.6925
      &0.8940 	&0.6337 	&0.4393 	&\textbf{0.9021} 	&0.5217 	&\multicolumn{1}{c|}{0.6782} 	&0.5767 	&0.3145 	&0.2309 	&\textbf{0.6458} 	&0.3474 	&0.4231\\
      U2Fusion \cite{xu2020u2fusion}	   &TPAMI'2020  &0.9179 &0.8661 &\multicolumn{1}{c|}{0.8920} &0.6488 &0.7137 &0.6813
      &0.8896 	&0.4174 	&0.4595 	&0.8933 	&0.4707 	&\multicolumn{1}{c|}{0.6261} 	&0.5464 	&0.2135 	&0.2384 	&0.6060 	&0.3061 	&0.3821\\
      DCTNet \cite{Li_DCTNet}	           &TIM'2023  &0.9266 &\underline{0.9143} &\multicolumn{1}{c|}{\underline{0.9204}} &0.6699 &\underline{0.7423} &\underline{0.7061}
      &0.8942 	&\underline{0.6484} 	&0.4911 	&0.8947 	&0.5749 	&\multicolumn{1}{c|}{\underline{0.7007}} 	&\underline{0.5776} 	&0.3377 	&\textbf{0.2810} 	&0.6415 	&0.3842 	&\underline{0.4444}\\
      DATFuse \cite{tang2023datfuse} 	   &TCSVT'2023  &0.9065 &0.8470 &\multicolumn{1}{c|}{0.8767} &0.6487 &0.6925 &0.6706
      &0.8888 	&0.5285 	&0.4586 	&0.8841 	&0.4439 	&\multicolumn{1}{c|}{0.6408} 	&0.5350 	&0.2738 	&0.2456 	&0.5891 	&0.2885 	&0.3864\\
      CrossFuse \cite{Wang_CrossFuse}	   &TCSVT'2023  &0.8033 &0.8436 &\multicolumn{1}{c|}{0.8235} &0.4521 &0.6637 &0.5579
      &0.8864 	&0.4991 	&0.4618 	&0.8894 	&0.4038 	&\multicolumn{1}{c|}{0.6281} 	&0.5522 	&0.2631 	&0.2523 	&0.5764 	&0.2467 	&0.3781\\
      SegMiF \cite{liu2023segmif}        &ICCV'2023   &0.9168 &0.8640 &\multicolumn{1}{c|}{0.8904} &0.6643 &0.7103 &0.6873
      &0.8922 	&0.4594 	&0.4741 	&0.8921 	&0.5842 	&\multicolumn{1}{c|}{0.6604} 	&0.5634 	&0.2631 	&0.2580 	&0.6173 	&\underline{0.3901} 	&0.4184\\
      LRRNet \cite{li2023lrrnet}         &TPAMI'2023  &0.8574 &0.8750 &\multicolumn{1}{c|}{0.8662} &0.5861 &0.7250 &0.6555
      &0.8926 	&0.5354 	&\underline{0.5061} 	&0.8970 	&0.5187 	&\multicolumn{1}{c|}{0.6700} 	&0.5640 	&0.3025 	&0.2702 	&0.6389 	&0.3468 	&0.4245\\
      DDFM    \cite{2023DDFM}            &ICCV'2023   &0.4822 &0.4594 &\multicolumn{1}{c|}{0.4708} &0.3397 &0.3700 &0.3548
      &0.8816 	&0.4704 	&0.4798 	&0.8774 	&0.3809 	&\multicolumn{1}{c|}{0.6180} 	&0.5300 	&0.2550 	&0.2305 	&0.5696 	&0.2507 	&0.3671\\
      SHIP    \cite{2024ship}            &CVPR'2024   &0.8983 &0.8402 &\multicolumn{1}{c|}{0.8692} &0.6604 &0.7133 &0.6868
      &\underline{0.8949} 	&0.6228 	&0.4844 	&0.8984 	&0.5238 	&\multicolumn{1}{c|}{0.6849} 	&0.5712 	&0.3204 	&0.2598 	&0.6412 	&0.3564 	&0.4298\\  
      MoCTEFuse	                 &- &0.9360 &\textbf{0.9200} &\multicolumn{1}{c|}{\textbf{0.9280}} &0.6459 &\textbf{0.7727} &\textbf{0.7093}
      &\textbf{0.8963} 	&\textbf{0.6573} 	&\textbf{0.5502} 	&\underline{0.8986} 	&\textbf{0.5949} 	&\multicolumn{1}{c|}{\textbf{0.7194}} 	&\textbf{0.5794} 	&\textbf{0.3625} 	&\underline{0.2793} 	&\underline{0.6443} 	&\textbf{0.3914} 	&\textbf{0.4514}\\
    \bottomrule
    \end{tabular}
    }
    \end{table*}
\subsubsection{Quantitative Comparison}
  \par{
    We further select 21 groups of source images from the TNO dataset and 221 sets of source images from the RoadScene dataset for quantitative assessment. The statistical results of thirteen representative fusion methods on the four metrics are shown in \autoref{table:MSRS}, \autoref{fig:allimgs}\subref{fig:c}, and \autoref{fig:allimgs}\subref{fig:d}. As listed in \autoref{table:MSRS}, our proposed MoCTEFuse achieves the highest mean values for the SD, MI, and VIF metrics in the TNO dataset. Similarly, in the RoadScene dataset, it outperforms other methods in all four metrics. These results suggest that MoCTEFuse effectively enhances contrast, retains more information from source images in fused images, and generates fused results that align with the human visual system. Besides, we observe that MoCTEFuse ranks in the middle range, slightly behiend DCTNet, TarDAL, and SegMiF in terms of the EN metric in the TNO dataset. This is attributed to the fact that the TNO dataset mainly consists of high-illumination scenes, where IR images are often heavily disturbed by noise. The influence of noise interference on EN metric is bidirectional. On the one hand, noise can increase the information complexity of images and thus elevate the EN value (\textit{e.g.}, DCTNet, TarDAL, and SegMiF). On the other hand, excessive or improperly handled noise may mask crucial information and cause uneven distribution of gray levels and thus reduce the EN values (\textit{e.g.}, U2Fusion and SHIP).
    
    In summary, both qualitative and quantitative results consistently demonstrate the superior generalization capabilities of MoCTEFuse. This benefits from the utilization of the competitive learning loss function and the MoCTE.
  }
    \subsection{Extension to Object Detection}
    It is well-known that an effective fusion method should not only generate visually appealing fused images but also enhance the performance of high-level computer vision tasks. This section broadens the analysis to general object detection and oriented object detection by utilizing single-modal images as well as fused images. The mean average precision (mAP) is employed as the main evaluation metric for this experiment. The detection results, specifically the average precision (AP) at IoU thresholds of 0.5 (\textit{i.e.}, AP0.5) and from 0.5 to 0.95 (\textit{i.e.}, AP0.5:0.95), for images before and after the application of state-of-the-art fusion techniques are presented in \autoref{table:detection}.
    
    \subsubsection{General Object Detection}
    We randomly select and manually annotate 80 pairs of typical images from the MFNet dataset \cite{ha2017mfnet}. To mitigate potential biases from fine-tuning, we directly input IR, VI, and fused images into a pre-trained model to assess their detection capabilities. For this study, we employ the YOLOX-s \cite{yolox2021} as the detector. 
    
    As evident from \autoref{table:detection}, our MoCTEFuse achieves AP0.5 of 92.80\% and AP0.5:0.95 of 70.93\%, significantly outperforming other methods. Additionally, the visualizations of the detection results in \autoref{fig:mfnet}, demonstrate that MoCTEFuse excels in minimizing missed detections and accurately locating salient objects. The detections are complete with precise bounding boxes and satisfactory confidence scores. This indicates that the proposed MoCTEFuse robustly enhances detection performance across various challenging environments.

    \subsubsection{Oriented Object Detection}
    The state-of-the-arts described in \autoref{methods} are employed to fuse VI and IR image pairs, producing corresponding fused images. Then, we utilize a widely used oriented detector \cite{2023psc} to train and test these images. Specifically, we choose a version of yolov5s combined with a dual-frequency phase-shifting coder (PSCD) because of its efficiency and outstanding detection performance. Following the same configurations as MMRotate \cite{zhou2022mmrotate}, we implement and evaluate all algorithms within a unified code repository.

    From \autoref{table:detection}, it is evident that the fused images generated by the proposed MoCTEFuse significantly outperform those produced by other approaches on the DroneVehicle dataset across AP0.5 and AP0.5:0.95. Specifically, the AP0.5/AP0.5:0.95 values for VI images are 0.3666/0.2236, while for IR images, they are 0.5445/0.3132. After applying MoCTEFuse, the detection accuracies increase by 0.3528/0.2278 and 0.1749/0.1382, respectively. \autoref{fig:drones} illustrates two pairs of detection results on the DroneVehicle dataset. Notably, in \autoref{fig:drones}\subref{fig:droneD}, a false detection occurs in the IR image, and a detection is missed in the SegMiF image. Additionally, the confidence scores of other methods are generally lower than those of our MoCTEFuse. In \autoref{fig:drones}\subref{fig:droneN}, missing detections are observed in the IR and DATFuse images, along with false detections in the VI, SwinFusion, YDTR, TarDAL, U2Fusion, DCTNet, CrossFuse, SegMiF, LRRNet, and SHIP images (\textit{e.g.}, a bus is misidentified as a truck). Compared to PIAFusion and DDFM, MoCTEFuse achieves more precise bounding boxes and higher confidence scores.
    
\section{Conclusion}\label{section5}
  In this study, we delve into the significance of illumination changes in image fusion, aiming to devise a robust fusion algorithm for infrared and visible images. Thus, we have proposed a dynamic fusion framework, called MoCTEFuse, which exhibits a precise perception and response to illumination interferences. By predicting the illumination distribution of visible images, the proposed MoCTE performs scene-adaptive sample allocation and processing. The core of MoCTE is to design the CTFB, which leverages an asymmetric cross-attention mechanism to progressively aggregate and refine cross-modality features. Furthermore, the proposed loss function incorporates SSIM, intensity, gradient, and illumination distribution into the network training, ensuring that the fused image retains crucial information from the source images while embodying attractive visual appearance. This work provides a valuable reference for intelligent perception systems, particularly involving illumination variations or Transformers. While MoCTEFuse achieves impressive fusion and detection performance, it also has a relatively high computational cost. In future work, we would like to design a lightweight method.

\bibliographystyle{IEEEtran}
\bibliography{fuse-refs}

\begin{thebibliography}{10}
\providecommand{\url}[1]{#1}
\csname url@samestyle\endcsname
\providecommand{\newblock}{\relax}
\providecommand{\bibinfo}[2]{#2}
\providecommand{\BIBentrySTDinterwordspacing}{\spaceskip=0pt\relax}
\providecommand{\BIBentryALTinterwordstretchfactor}{4}
\providecommand{\BIBentryALTinterwordspacing}{\spaceskip=\fontdimen2\font plus
\BIBentryALTinterwordstretchfactor\fontdimen3\font minus \fontdimen4\font\relax}
\providecommand{\BIBforeignlanguage}[2]{{%
\expandafter\ifx\csname l@#1\endcsname\relax
\typeout{** WARNING: IEEEtran.bst: No hyphenation pattern has been}%
\typeout{** loaded for the language `#1'. Using the pattern for}%
\typeout{** the default language instead.}%
\else
\language=\csname l@#1\endcsname
\fi
#2}}
\providecommand{\BIBdecl}{\relax}
\BIBdecl

\bibitem{2024OAFA}
C.~Chen, J.~Qi, X.~Liu, K.~Bin, R.~Fu, X.~Hu, and P.~Zhong, ``Weakly misalignment-free adaptive feature alignment for uavs-based multimodal object detection,'' in \emph{Proc. IEEE/CVF Conf. Comput. Vis. Pattern Recognit., CVPR}, 2024, pp. 26\,826--26\,835.

\bibitem{huang2023multi}
Z.~Huang, S.~Sun, J.~Zhao, and L.~Mao, ``Multi-modal policy fusion for end-to-end autonomous driving,'' \emph{Inf. Fusion}, vol.~98, p. 101834, 2023.

\bibitem{2023robot}
\BIBentryALTinterwordspacing
Q.~Gao, Z.~Deng, Z.~Ju, and T.~Zhang, ``Dual-hand motion capture by using biological inspiration for bionic bimanual robot teleoperation,'' \emph{Cyborg and Bionic Systems}, vol.~4, p. 0052, 2023. [Online]. Available: \url{https://spj.science.org/doi/abs/10.34133/cbsystems.0052}
\BIBentrySTDinterwordspacing

\bibitem{hill2016perceptual}
P.~Hill, M.~E. Al-Mualla, and D.~Bull, ``Perceptual image fusion using wavelets,'' \emph{IEEE Trans. Image Process.}, vol.~26, no.~3, pp. 1076--1088, 2017.

\bibitem{2010MCT}
X.~Chang, L.~Jiao, F.~Liu, and F.~Xin, ``Multicontourlet-based adaptive fusion of infrared and visible remote sensing images,'' \emph{IEEE Geosci. Remote Sens. Lett.}, vol.~7, no.~3, pp. 549--553, 2010.

\bibitem{2020MDLatLRR}
H.~Li, X.-J. Wu, and J.~Kittler, ``Mdlatlrr: A novel decomposition method for infrared and visible image fusion,'' \emph{IEEE Trans. Image Process.}, vol.~29, pp. 4733--4746, 2020.

\bibitem{cvejic2007region}
N.~Cvejic, D.~Bull, and N.~Canagarajah, ``Region-based multimodal image fusion using ica bases,'' \emph{IEEE Sens. J.}, vol.~7, no.~5, pp. 743--751, 2007.

\bibitem{liu2015general}
Y.~Liu, S.~Liu, and Z.~Wang, ``A general framework for image fusion based on multi-scale transform and sparse representation,'' \emph{Inf. Fusion}, vol.~24, pp. 147--164, 2015.

\bibitem{dogra2017multi}
A.~Dogra, B.~Goyal, and S.~Agrawal, ``From multi-scale decomposition to non-multi-scale decomposition methods: a comprehensive survey of image fusion techniques and its applications,'' \emph{IEEE Access}, vol.~5, pp. 16\,040--16\,067, 2017.

\bibitem{zhu2017fusion}
P.~Zhu, X.~Ma, and Z.~Huang, ``Fusion of infrared-visible images using improved multi-scale top-hat transform and suitable fusion rules,'' \emph{Infrared Phys. Technol.}, vol.~81, pp. 282--295, 2017.

\bibitem{xu2020u2fusion}
H.~Xu, J.~Ma, J.~Jiang, X.~Guo, and H.~Ling, ``U2fusion: A unified unsupervised image fusion network,'' \emph{IEEE Trans. Pattern Anal. Mach. Intell.}, vol.~44, no.~1, pp. 502--518, 2020.

\bibitem{2024ship}
N.~Zheng, M.~Zhou, J.~Huang, J.~Hou, H.~Li, Y.~Xu, and F.~Zhao, ``Probing synergistic high-order interaction in infrared and visible image fusion,'' in \emph{Proc. IEEE/CVF Conf. Comput. Vis. Pattern Recognit., CVPR}, 2024, pp. 26\,384--26\,395.

\bibitem{2022TarDAL}
J.~Liu, X.~Fan, Z.~Huang, G.~Wu, R.~Liu, W.~Zhong, and Z.~Luo, ``Target-aware dual adversarial learning and a multi-scenario multi-modality benchmark to fuse infrared and visible for object detection,'' in \emph{Proc. IEEE/CVF Conf. Comput. Vis. Pattern Recognit., CVPR}, 2022, pp. 5802--5811.

\bibitem{Wang_CrossFuse}
Z.~Wang, W.~Shao, Y.~Chen, J.~Xu, and L.~Zhang, ``A cross-scale iterative attentional adversarial fusion network for infrared and visible images,'' \emph{IEEE Trans. Circuits Syst. Video Technol.}, vol.~33, no.~8, pp. 3677--3688, 2023.

\bibitem{2022SwinFusion}
J.~Ma, L.~Tang, F.~Fan, J.~Huang, X.~Mei, and Y.~Ma, ``Swinfusion: Cross-domain long-range learning for general image fusion via swin transformer,'' \emph{IEEE-CAA J. Automatica Sin.}, vol.~9, no.~7, pp. 1200--1217, 2022.

\bibitem{LImixfuse}
\BIBentryALTinterwordspacing
J.~Li, H.~Song, L.~Liu, Y.~Li, J.~Xia, Y.~Huang, J.~Fan, Y.~Lin, and J.~Yang, ``Mixfuse: An iterative mix-attention transformer for multi-modal image fusion,'' \emph{Expert Systems with Applications}, vol. 261, p. 125427, 2025. [Online]. Available: \url{https://www.sciencedirect.com/science/article/pii/S0957417424022942}
\BIBentrySTDinterwordspacing

\bibitem{2023DDFM}
Z.~Zhao, H.~Bai, Y.~Zhu, J.~Zhang, S.~Xu, Y.~Zhang, K.~Zhang, D.~Meng, R.~Timofte, and L.~Van~Gool, ``Ddfm: Denoising diffusion model for multi-modality image fusion,'' in \emph{Proc. IEEE/CVF Int. Conf. Comput. Vis., ICCV}, 2023, pp. 8048--8059.

\bibitem{tang2022ydtr}
W.~Tang, F.~He, and Y.~Liu, ``Ydtr: infrared and visible image fusion via y-shape dynamic transformer,'' \emph{IEEE Trans. Multimedia}, 2022.

\bibitem{li2023lrrnet}
H.~Li, T.~Xu, X.-J. Wu, J.~Lu, and J.~Kittler, ``Lrrnet: A novel representation learning guided fusion network for infrared and visible images,'' \emph{IEEE Trans. Pattern Anal. Mach. Intell.}, vol.~45, no.~9, pp. 11\,040--11\,052, 2023.

\bibitem{2019Tong}
Y.~Tong and J.~Chen, ``Infrared and visible image fusion under different illumination conditions based on illumination effective region map,'' \emph{IEEE Access}, vol.~7, pp. 151\,661--151\,668, 2019.

\bibitem{zhang2024evfusion}
X.~Zhang, X.~Wang, C.~Yan, and Q.~Sun, ``Ev-fusion: A novel infrared and low-light color visible image fusion network integrating unsupervised visible image enhancement,'' \emph{IEEE Sens. J.}, 2024.

\bibitem{yang2024iaifnet}
Q.~Yang, Y.~Zhang, Z.~Zhao, J.~Zhang, and S.~Zhang, ``Iaifnet: An illumination-aware infrared and visible image fusion network,'' \emph{IEEE Signal Process. Lett.}, 2024.

\bibitem{LUO2024FVC}
Y.~Luo, D.~Xu, K.~He, H.~Shi, and J.~Gong, ``Fidelity based visual compensation and salient information rectification for infrared and visible image fusion,'' \emph{Knowledge-Based Syst.}, vol. 299, p. 112132, 2024.

\bibitem{tang2022piafusion}
L.~Tang, J.~Yuan, H.~Zhang, X.~Jiang, and J.~Ma, ``Piafusion: A progressive infrared and visible image fusion network based on illumination aware,'' \emph{Inf. Fusion}, vol.~83, pp. 79--92, 2022.

\bibitem{cao2023moefusion}
B.~Cao, Y.~Sun, P.~Zhu, and Q.~Hu, ``Multi-modal gated mixture of local-to-global experts for dynamic image fusion,'' in \emph{Proc. IEEE/CVF Int. Conf. Comput. Vis., ICCV}, 2023, pp. 23\,555--23\,564.

\bibitem{liu2018infrared}
Y.~Liu, X.~Chen, J.~Cheng, H.~Peng, and Z.~Wang, ``Infrared and visible image fusion with convolutional neural networks,'' \emph{Int. J. Wavelets Multiresolut. Inf. Process.}, vol.~16, no.~03, p. 1850018, 2018.

\bibitem{RN25}
H.~Li and X.-J. Wu, ``Densefuse: A fusion approach to infrared and visible images,'' \emph{IEEE Trans. Image Process.}, vol.~28, no.~5, pp. 2614--2623, 2019.

\bibitem{2019FusionGAN}
J.~Ma, W.~Yu, P.~Liang, C.~Li, and J.~Jiang, ``Fusiongan: A generative adversarial network for infrared and visible image fusion,'' \emph{Inf. Fusion}, vol.~48, pp. 11--26, 2019.

\bibitem{tang2022seafusion}
L.~Tang, J.~Yuan, and J.~Ma, ``Image fusion in the loop of high-level vision tasks: A semantic-aware real-time infrared and visible image fusion network,'' \emph{Inf. Fusion}, vol.~82, pp. 28--42, 2022.

\bibitem{zhao2023cddfuse}
Z.~Zhao, H.~Bai, J.~Zhang, Y.~Zhang, S.~Xu, Z.~Lin, R.~Timofte, and L.~Van~Gool, ``Cddfuse: Correlation-driven dual-branch feature decomposition for multi-modality image fusion,'' in \emph{Proc. IEEE/CVF Conf. Comput. Vis. Pattern Recognit., CVPR}, 2023, pp. 5906--5916.

\bibitem{MBNet-ECCV2020}
K.~Zhou, L.~Chen, and X.~Cao, ``Improving multispectral pedestrian detection by addressing modality imbalance problems,'' in \emph{Proc. Eur. Conf. Comput. Vis., ECCV}, 2020.

\bibitem{2023CSIM}
Y.-F. Lu, J.-W. Gao, Q.~Yu, Y.~Li, Y.-S. Lv, and H.~Qiao, ``A cross-scale and illumination invariance-based model for robust object detection in traffic surveillance scenarios,'' \emph{IEEE Trans. Intell. Transp. Syst.}, vol.~24, no.~7, pp. 6989--6999, 2023.

\bibitem{Guo_2020_CVPR}
C.~Guo, C.~Li, J.~Guo, C.~C. Loy, J.~Hou, S.~Kwong, and R.~Cong, ``Zero-reference deep curve estimation for low-light image enhancement,'' in \emph{Proc. IEEE/CVF Conf. Comput. Vis. Pattern Recognit., CVPR}, 2020.

\bibitem{Wang_2023_ICCV}
Y.~Wang, Z.~Liu, J.~Liu, S.~Xu, and S.~Liu, ``Low-light image enhancement with illumination-aware gamma correction and complete image modelling network,'' in \emph{Proc. IEEE/CVF Int. Conf. Comput. Vis., ICCV}, Oct 2023, pp. 13\,128--13\,137.

\bibitem{6797059}
R.~A. Jacobs, M.~I. Jordan, S.~J. Nowlan, and G.~E. Hinton, ``Adaptive mixtures of local experts,'' \emph{Neural Computation}, vol.~3, no.~1, pp. 79--87, 1991.

\bibitem{716791}
M.~Jordan and R.~Jacobs, ``Hierarchical mixtures of experts and the em algorithm,'' in \emph{Proc. Int. Jt. Conf. Neural. Networks.}, vol.~2, 1993, pp. 1339--1344.

\bibitem{2021Gshard}
D.~Lepikhin, H.~Lee, Y.~Xu, D.~Chen, O.~Firat, Y.~Huang, M.~Krikun, N.~Shazeer, and Z.~Chen, ``Gshard: Scaling giant models with conditional computation and automatic sharding,'' in \emph{Int. Conf. Learn. Represent., ICLR}, 2021.

\bibitem{xue2022go}
F.~Xue, Z.~Shi, F.~Wei, Y.~Lou, Y.~Liu, and Y.~You, ``Go wider instead of deeper,'' in \emph{Proc. AAAI Conf. Artif. Intell., AAAI}, vol.~36, 2022, pp. 8779--8787.

\bibitem{NEURIPS2021_VMoE}
C.~Riquelme, J.~Puigcerver, B.~Mustafa, M.~Neumann, R.~Jenatton, A.~Susano~Pinto, D.~Keysers, and N.~Houlsby, ``Scaling vision with sparse mixture of experts,'' in \emph{Adv. neural inf. proces. syst., NeurIPS}, vol.~34, 2021, pp. 8583--8595.

\bibitem{chen2023mod}
Z.~Chen, Y.~Shen, M.~Ding, Z.~Chen, H.~Zhao, E.~G. Learned-Miller, and C.~Gan, ``Mod-squad: Designing mixtures of experts as modular multi-task learners,'' in \emph{Proc. IEEE/CVF Conf. Comput. Vis. Pattern Recognit., CVPR}, 2023, pp. 11\,828--11\,837.

\bibitem{zhao2024removal}
T.~Zhao, M.~Yuan, and X.~Wei, ``Removal and selection: Improving rgb-infrared object detection via coarse-to-fine fusion,'' \emph{arXiv:2401.10731}, 2024.

\bibitem{Li_DCTNet}
J.~Li, L.~Liu, H.~Song, Y.~Huang, J.~Jiang, and J.~Yang, ``Dctnet: A heterogeneous dual-branch multi-cascade network for infrared and visible image fusion,'' \emph{IEEE Trans. Instrum. Meas.}, vol.~72, pp. 1--14, 2023.

\bibitem{resnet}
K.~He, X.~Zhang, S.~Ren, and J.~Sun, ``Deep residual learning for image recognition,'' in \emph{Proc. IEEE/CVF Conf. Comput. Vis. Pattern Recognit., CVPR}, 2016, pp. 770--778.

\bibitem{sun2020drone}
Y.~Sun, B.~Cao, P.~Zhu, and Q.~Hu, ``Drone-based rgb-infrared cross-modality vehicle detection via uncertainty-aware learning,'' \emph{IEEE Trans. Circuits Syst. Video Technol.}, pp. 1--1, 2022.

\bibitem{tang2023datfuse}
W.~Tang, F.~He, Y.~Liu, Y.~Duan, and T.~Si, ``Datfuse: Infrared and visible image fusion via dual attention transformer,'' \emph{IEEE Trans. Circuits Syst. Video Technol.}, 2023.

\bibitem{liu2023segmif}
J.~Liu, Z.~Liu, G.~Wu, L.~Ma, R.~Liu, W.~Zhong, Z.~Luo, and X.~Fan, ``Multi-interactive feature learning and a full-time multi-modality benchmark for image fusion and segmentation,'' in \emph{Proc. IEEE/CVF Int. Conf. Comput. Vis., ICCV}, 2023.

\bibitem{ha2017mfnet}
Q.~Ha, K.~Watanabe, T.~Karasawa, Y.~Ushiku, and T.~Harada, ``Mfnet: Towards real-time semantic segmentation for autonomous vehicles with multi-spectral scenes,'' in \emph{Proc. IEEE/RSJ Int. Conf. Intell. Robots Syst., IROS}, 2017, pp. 5108--5115.

\bibitem{yolox2021}
Z.~Ge, S.~Liu, F.~Wang, Z.~Li, and J.~Sun, ``Yolox: Exceeding yolo series in 2021,'' \emph{arXiv:2107.08430}, 2021.

\bibitem{2023psc}
Y.~Yu and F.~Da, ``Phase-shifting coder: Predicting accurate orientation in oriented object detection,'' in \emph{Proc. IEEE/CVF Conf. Comput. Vis. Pattern Recognit., CVPR}, 2023, pp. 13\,354--13\,363.

\bibitem{zhou2022mmrotate}
Y.~Zhou, X.~Yang, G.~Zhang, J.~Wang, Y.~Liu, L.~Hou, X.~Jiang, X.~Liu, J.~Yan, C.~Lyu, W.~Zhang, and K.~Chen, ``Mmrotate: A rotated object detection benchmark using pytorch,'' in \emph{MM - Proc. ACM Int. Conf. Multimed.}, 2022.

\end{thebibliography}
\end{document}